\documentclass[sigconf]{acmart}

\usepackage{microtype}
\usepackage{graphicx}
\usepackage{subfigure}
\usepackage{booktabs} % for professional tables
\usepackage{amsmath}
\usepackage{textcomp}
\usepackage{xcolor}
\usepackage{verbatim,color}
\usepackage{xcolor}
\usepackage{listings}
\usepackage{color}
\usepackage{CJK}
\usepackage{framed}
\usepackage{natbib}
\usepackage{chngcntr}
\usepackage{listings}
\usepackage{hyperref}

\renewcommand{\mathbf}[1]{{\boldsymbol #1}}
\hypersetup{
	colorlinks   = true,
	citecolor    = blue,
	linkcolor    = blue,
}
\graphicspath{{figures/}}

\AtBeginDocument{%
  \providecommand\BibTeX{{%
    \normalfont B\kern-0.5em{\scshape i\kern-0.25em b}\kern-0.8em\TeX}}}

\setcopyright{acmcopyright}
\copyrightyear{2020}
\acmYear{2020}
\acmDOI{10.1145/AAA.BBB}

\acmConference[Technical Report]{}
\acmBooktitle{May 2020}
\acmPrice{15.00}
\acmISBN{AAA-B-CCCC-DDDD-E/FF/GG}

\begin{document}

%%
%% The "title" command has an optional parameter,
%% allowing the author to define a "short title" to be used in page headers.
\title{Understanding the Nature of System-Related Issues in Machine Learning Frameworks: An Exploratory Study}

%%
%% The "author" command and its associated commands are used to define
%% the authors and their affiliations.
%% Of note is the shared affiliation of the first two authors, and the
%% "authornote" and "authornotemark" commands
%% used to denote shared contribution to the research.

 % \author{Anonymous}
 % \affiliation{%
 %   \institution{University of Anonymous}
 %   \city{Somewhere}
 %   \country{Some Country}}
 %  \email{anonymous@anonymous.edu}

\author{Yang Ren}
\affiliation{%
  \institution{University of South Carolina}
%   \streetaddress{1 Th{\o}rv{\"a}ld Circle}
%   \city{Hekla}
  \country{USA}}
% \email{larst@affiliation.org}

\author{Gregory Gay}
\affiliation{%
  \institution{Chalmers and the University of Gutenberg}
%   \streetaddress{1 Th{\o}rv{\"a}ld Circle}
%   \city{Hekla}
  \country{Sweden}}
% \email{larst@affiliation.org}

\author{Christian K{\"a}stner}
\affiliation{%
  \institution{Carnegie Mellon University}
%   \streetaddress{1 Th{\o}rv{\"a}ld Circle}
%   \city{Hekla}
  \country{USA}}
% \email{larst@affiliation.org}

\author{Pooyan Jamshidi}
\affiliation{%
  \institution{University of South Carolina}
%   \streetaddress{1 Th{\o}rv{\"a}ld Circle}
%   \city{Hekla}
  \country{USA}}
% \email{larst@affiliation.org}

%%
%% By default, the full list of authors will be used in the page
%% headers. Often, this list is too long, and will overlap
%% other information printed in the page headers. This command allows
%% the author to define a more concise list
%% of authors' names for this purpose.
\renewcommand{\shortauthors}{Yang Ren, Gregory Gay, Christian K{\"a}stner, Pooyan Jamshidi}

%%
%% The abstract is a short summary of the work to be presented in the
%% article.
\begin{abstract}
\noindent \textbf{Background:} Modern systems are built using \textit{development frameworks}. These frameworks have a major impact on how the resulting system executes, how configurations are managed, how it is tested, and how and where it is deployed. Machine learning (ML) frameworks and the systems developed using them differ greatly from traditional frameworks. Naturally, the issues that manifest in such frameworks may differ as well---as may the behavior of developers addressing those issues.

\noindent \textbf{Aims:} We are interested in characterizing the system-related issues---issues impacting performance, memory and resource usage, and other quality attributes---that emerge in ML frameworks, and how they differ from those in traditional frameworks. 

\noindent \textbf{Method:} We have conducted a moderate-scale exploratory study analyzing real-world system-related issues from 10 popular machine learning frameworks.

\noindent \textbf{Results:} Our findings offer implications for the development of machine learning systems, including differences in the frequency of occurrence of certain issue types, observations regarding the impact of debate and time on issue correction, and differences in the specialization of developers.

\noindent \textbf{Conclusions:} We hope that this exploratory study will enable developers to improve their expectations, plan for risk, and allocate resources accordingly when making use of the tools provided by these frameworks to develop ML-based systems.
\end{abstract}

%%
%% The code below is generated by the tool at http://dl.acm.org/ccs.cfm.
%% Please copy and paste the code instead of the example below.
%%


%%
%% Keywords. The author(s) should pick words that accurately describe
%% the work being presented. Separate the keywords with commas.

\maketitle

\section{Introduction}
\label{Introduction}

A new paradigm of software systems have emerged, called machine learning (ML) systems~\cite{carbin2019overparameterization,k2017software2}. Traditional software systems consist of explicit instructions to a computer written by the programmer, whereas ML systems
learn behavior from data. These systems are designed as a skeletal code architecture---specifying high-level behavioral goals---layered over highly-optimized models~\cite{k2017software2}. ML systems have revolutionized business intelligence, health care, finance, and other industries that power society. 

Modern systems, both traditional and ML-based, are often powered by underlying \textit{frameworks}---libraries of services that are used for providing higher-level functionality. TensorFlow, for example, offers a library for developing, training, or deploying models for use in ML systems. In traditional domains, we can look to examples like React or Flutter---frameworks for building user interfaces---or Rancher---a framework that provides container services.

Some argue that \textit{best practices} for the development and quality assurance of traditional software systems are still largely based on ad-hoc experience, and often more closely represent an art than an established science~\cite{McConnell98:Art}. ML systems and frameworks are so different that many of the lessons learned from traditional software development may no longer apply. ML systems differ in how they are developed, how they execute, how configurations are managed, how systems are tested, and how and where those systems are deployed. Naturally, then, the faults that developers create and the failures that manifest as a result may differ as well---as may how communities of developers behave in correcting those issues. We expect many differences and that those differences can be attributed to the underlying frameworks powering such systems. Therefore, we wish to better understand the types of issues that tend to occur in machine learning frameworks, and how they compare and contrast to the issues that impact frameworks in traditional paradigms. 

Specifically, we are interested in characterizing and contrasting the types of \textit{system-related issues} that emerge in the infrastructure code provided by ML frameworks (e.g., TensorFlow) versus frameworks for more traditional tasks (i.e., React). With system-related issues, we refer to issues affecting quality attributes, such as performance, configuration, component interaction, and memory usage. System-related issues are common in all types of systems~\cite{sculley2015hidden}, but it is not yet understood how they impact ML frameworks---and the systems built using these frameworks---and how the characteristics of such issues differ between system paradigms.
Based on our experience in developing several ML and non-ML systems as well as systematic mining of issues in open source frameworks in GitHub, we hypothesize that the system-related issues in ML frameworks differ from traditional software frameworks in three key areas: (1) \textit{their types and frequencies of occurrence}, (2) \textit{the difficulty of fixing particular issue types}, and (3) \textit{the composition and behavior of the teams of developers engaged in fixing these issues}. 
% \todo{why do we expect this? the intro would be stronger if we could give some clearer reasons why we expect differences and why they matter, not just saying that something may be different.}
% In order to improve the engineering of ML-powered systems, it is important that we understand the types of issues likely to emerge from their underlying frameworks, how the shift in paradigm impacts the difficulty of addressing issues, and how community behavior evolves during the process of fixing such issues. 
A better understanding of system-related issues across domains will enable developers to improve their expectations, plan for risk, and better allocate resources.

In order to explore this topic, we have conducted an \textit{exploratory study of bug reports} in ML and non-ML frameworks. Guided by a number of research questions, we compare data gathered from ML frameworks with data from traditional software frameworks. Specifically, we manually analyzed and classified \textbf{121} system-related issues from 10 popular ML frameworks and a further \textbf{332} system-related issues from 10 traditional software frameworks, collected from the GitHub project of each system. 
%We manually classified these issues into different issue categories---listed in Table~\ref{table:category1}---such as configuration errors, performance regression, or memory leak. We then compared the distribution of issue types between categories of systems in order to better understand how the types of issues and frequency of occurrence of those types differs between paradigms. 
Furthermore, we approximate the difficulty of fixing an issue, including time, discussion, and patch size indicators, and examined the behavior of the community, as it relates to correcting issues---including the number of people involved in issue discussion and the level of activity involved in issue correction, among others. We explore the collected data for trends and differences.
Our findings include a number of observations:
% , with implications for the development of Software 2.0 systems:
\begin{itemize}
\setlength{\itemsep}{1pt}
  \setlength{\parskip}{0pt}
  \setlength{\parsep}{0pt}  
    \item  Incorrect memory allocation, memory leaks, multi-threading errors, and performance regression occur more commonly in ML frameworks---possibly due to the need to manage large quantities of data in parallel and the rapid pace of system enhancement. Increased dependence on hardware selection, like the GPU, can also lead to issues. Configuration errors are very common in traditional software frameworks, but rarely occur in ML frameworks. %, possibly because such systems tend to offer fewer explicit configurations.
    % \todo{import: avoid causal languages. just because you find a difference, doesn't mean there is causation. carefully state facts and separate possible explanations (speculation)}
    \item Issues in ML do not appear to be significantly more difficult to address. API mismatch issues require significant time and discussion to fix, reflecting rapidly evolving communities debating how to best evolve their systems. Incorrect memory allocation issues also attract significantly more participants in discussing potential fixes. The most contentious issues reflect an evolving field and an active community. Memory leaks attract less participation at the pull request level, indicating an area where issue commonality leads to quick acceptance of solutions.
    \item Users of ML frameworks provide more detailed issue descriptions. Issues are not necessarily harder to reproduce. Instead, users may be more knowledgeable, have more development experience, and may be more prepared to offer background on the issue being reported than in traditional frameworks.
    \item Many ML frameworks developers identify as a combination of Engineer and Researcher, while many traditional framework developers identify solely as an Engineer. ML framework developers also tend to be more popular than the developers of traditional frameworks. There is little consistency in how long developers have had GitHub accounts.
    \item There is little we can say categorically about community activity level for ML versus software frameworks.
     ML issues do not attract significantly more non-developer users to take part in discussion than software frameworks. Overall, the two categories show similar levels of member participation. Software frameworks members contribute more to open source software. However, there are significant differences between individual systems. 
\end{itemize}

To summarize, we make the following contributions:
\begin{itemize}
    \item A moderate-scale exploratory study of system-related issues and their root causes as well as cross-comparison on ten widely used ML and traditional software frameworks.
    \item An in-depth analysis, characterization, and classification of \textbf{453} system-related issues and their related patches.
    \item A quantitative comparison of the difficulty of fixing issues, community behavior, and the issue fixing process.
    \item Actionable recommendations to the developers of ML frameworks, as well as systems that make use of these frameworks.
    \item A replication package\footnote{\url{https://doi.org/10.5281/zenodo.3786191}} containing all data gathered in the process of performing this study. We hope that this exploratory study will offer assistance to the developers and researchers building ML systems and forming the best practices for ML-based fields. 
\end{itemize}

% In addition to our observations, we offer a replication package\footnote{\url{www.github.com/anonymous}} containing all data gathered in the process of performing this study. We hope that this exploratory study will offer assistance to the developers and researchers building ML systems and forming the best practices for ML-based fields. 

% \todo{be explicit about intended contributions}

\begin{table}[!t]
\caption{Issue categories and their definitions.}
\label{table:category1}
\vspace{-10pt}
\begin{center}
\begin{scriptsize}
\begin{tabular}{p{1in}p{2in}}
\toprule
\textbf{Category (Short Title)} & \textbf{Definition}\\
\midrule
API Mismatch (API) & Change to API version or mixed usage of APIs leading to performance degradation.\\
Compilation Error (Compl) & Failure to compile the source code.\\
Configuration Error (Config) & Configuration settings lead to performance degradation or error.\\
Connection Error (Conn) & Unexpected or wrongly-formatted connection request leads to error.\\
Data Race (Race) & Two or more threads access the same memory location concurrently.\\
Execution Error (Exec) & Unexpected error leads to the execution process crashing. \\
Hardware-Architecture Mismatch (HA) & Unfit hardware architecture leads to performance degradation or compilation error.\\
Memory Allocation (MA) &  Memory allocation leads to performance degradation.\\
I/O Slowdown (I/O) & Issues with I/O processes lead to performance degradation.\\
Memory Leak (ML) & A failure in a program to release memory. \\
Model Conversion (Conv) & Performance degradation due to type conversion/cast.\\
Multi-Threading Error (MT) & Performance  degradation due to thread interaction. \\
Performance Regression (PR) & Performance degradation after a change to the system.\\
Slow Synchronization (SYNC) & Synchronization between components leads to performance degradation. \\
Unexpected Resource Usage (RU) & Unusual system resource usage or requests leading to error or performance degradation. \\
\bottomrule
\end{tabular}
\end{scriptsize}
\end{center}
\vspace{-10pt}
\end{table}

\section{System-Related Issues}\label{sec:sysrel}

% \greg{TODO: Where did these categories come from? Did you get them from somewhere, or is this from personal experience? Reviewers wanted to know how you came up with this set}

We define a \textit{system-related issue} as a fault in the software that impacts quality attributes (non-functional properties) of the system, rather than functional issues, which result in the software producing the incorrect output. System-related issues tend to lead to performance degradation, loss of security, inappropriate usage of disc resources, or reduction in service~\cite{gregg2013systems}. System-related issues are significant, as they are a critical in determining system reliability and user experience. They are also useful in characterizing categories of systems, as unlike functional issues, system-related issued are not typically tied to system-specific requirements~\cite{Bass12:SAP}.

In this study, we have manually classified sampled issues into fifteen categories. Those categories are listed in Table~\ref{table:category1}. We derived these categories through manual coding~\cite{stol2016grounded}. More specifically, we used open coding to transform the initial structure into unstructured text by abstracting from large amounts of textual descriptions of issues and assigning codes to single textual description. One of the authors read the description of a new issue and if there exist an existing code in the taxonomy, he will then assign it to the issue, otherwise, he would create a new code for the new issue. One of the other authors then review the codes and refine the name and check whether the assignment was done correctly. In case of disagreement, then they have discussed the details of each issue to come to a resolution by renaming the issue code, assigning to another category, or simply creating a new code for the issue.

%We do not claim that this is an exhaustive list of all types of issues. 
These categories reflect the root causes of all of the sampled issues. To help illustrate the core concept, we present here examples of system-related issues in the studied ML frameworks:

\noindent\textbf{Unexpected Resource Usage:} A PyTorch user complained of ``too many resources requested'' errors shortly after the release of JetPack 3.2\footnote{\scriptsize{\url{https://github.com/pytorch/pytorch/issues/7680}}}. The developers found that the compiler lacked knowledge of how many threads the user wished to launch with, and the kernel was compiled to request more registers than is available on NVIDIA TX2. The patch added launch bounds that point out the maximum number of threads, so the kernel would not overuse registers.
%In Pytorch GitHub repository, a user reported a runtime error about too many resources requested when launching CUDA. After the user reduced the size of the model, he still got the same error. This error happened basically on the JetPack 3.2 which is just released before this issue. Before NVIDIA release this version for TX2, this kind of issue never happened. After the user discusses with the developer of Pytorch, they found that the reason of this issue is compiler does not know how many threads you want to launch with and the kernel is compiled to request more registers than available on TX2.  Quickly, the Pytorch team release the patch to fix this issue: 
%\ref{appendix:a}
%In this patch, they add the launch bounds to im2col and col2im. In this way, the launch bound will point out the max number of threads, and the kernel will not overuse registers when launching the process.

\iffalse
\newcommand{\lstbg}[3][0pt]{{\fboxsep#1\colorbox{#2}{\strut #3}}}
\lstdefinelanguage{diff}{
  basicstyle=\ttfamily\scriptsize,
  morecomment=[f][\lstbg{red!20}]-,
  morecomment=[f][\lstbg{green!20}]+,
  morecomment=[f][\textit]{@@},
  %morecomment=[f][\textit]{---},
  %morecomment=[f][\textit]{+++},
}
\vspace{-10pt}
\begin{lstlisting}[language=diff]
template <typename Dtype>
+ __launch_bounds__(CUDA_NUM_THREADS)
__global__ void im2col_kernel(const int n, 
            const Dtype* data_im,
            const int height, const int width,
            const int ksize_h, const int ksize_w,
\end{lstlisting}
\fi

\noindent\textbf{Performance Regression:} A Keras user reported that version 2.0.9 was extremely slow compared to 2.0.8\footnote{\scriptsize{\url{https://github.com/keras-team/keras/issues/8381}}}. For example, training a model went from 1-2 seconds to 10+ seconds, despite no environmental changes. After examining a variety of component interactions, the developers discovered the source of the slowdown---a method counting individual parameters. A simple check on the number of weights could be used without impacting functional correctness of the code, and without incurring slowdown. 

\noindent\textbf{Memory Leak:} A TensorFlow user reported that a simple code fragment, creating a queue structure, would consume 10GB of memory\footnote{\scriptsize{\url{https://github.com/tensorflow/tensorflow/issues/2942}}}. A contributor found that the root cause was heap fragmentation, resulting from input being copied into new arrays on each step. This led to rapid changes to the memory heap, which were not handled well by \texttt{malloc}. The patch fixing this issue reduced the number of unnecessary array allocations by using a function that pulls values directly instead of copying them to a new array first. 
%\textbf{Memory Leak (Tensorflow)}: In Tensorflow GitHub repository, a user created a FIFO query for numpy arrays in Tensorflow basic on Python2+Linux+GPU environment. And the user found that the program consumes 10GB main memory quickly. Memory leak is a primary category of system related issue. After a contributor of Tensorflow investigated this issue, the user found the reason is that the incoming feed dict values copied into new numpy arrays on every step and heap fragmentation arising from the creation of a large number of numPy arrays. After that, Tensorflow team released a patch to fixed this issue quickly: \ref{appendix:a}\\
%In this patch, they use `numpy.asarray()` instead of `numpy.array()` when converting feeds. The main difference between ‘array’ and ‘asarray’ is that the ‘array’ will make a copy of the object on every step when converting feeds, but the ‘asarray’ will avoids the copy. In this way, it reduces the number of unnecessary NumPy array allocations when feeding values into a TensorFlow session, and reduce heap fragmentation.\\

\iffalse
\vspace{-10pt}
\begin{lstlisting}[language=diff]
- np_val = np.array(subfeed_val, 
-                   dtype=subfeed_dtype)
+ np_val = np.asarray(subfeed_val, 
+                   dtype=subfeed_dtype)
\end{lstlisting}
\fi

\noindent\textbf{API Mismatch:} A TensorFlow user reported a crash following the use of a method from the \texttt{dataset} API on a dataset containing nested elements\footnote{\scriptsize{\url{https://github.com/tensorflow/tensorflow/issues/17932}}}. The issue was with a function used to group input by variable length. The API was updated to correctly unpack input with nested arguments.
%\textbf{API Issue (Tensorflow)}: In Tensorflow GitHub repository, a user reported that the dataset API does not support the ‘bucket\_by\_sequence\_length’ function with tuple dataset elements. This is a new function to group variable length inputs for the dataset API when the issue happened. After the developers of the system notice and investigate this issue, they found that the new function applied with dataset API was not correctly handling tupleized elements. After that, the developers released a patch to fix this issue: \ref{appendix:a}\\
%This patch fixes the behavior of ‘bucket\_by\_sequence\_length’ function with tuple Dataset elements. The developer replaced ‘element’ with ‘*args’ of the ‘element\_to\_bucket\_id’ function in the python operations` grouping file. In this way, the ‘element\_to\_bucket\_id()’ function will not take exactly one argument when given two or more elements. \\
\iffalse
\vspace{-10pt}
\begin{lstlisting}[language=diff]
' Try explicitly setting the type of the feed tensor'
' to a larger type (e.g. int64).')
- def element_to_bucket_id(element):
+ def element_to_bucket_id(*args):
- seq_length = element_length_func(element)
+ seq_length = element_length_func(*args)
boundaries = list(bucket_boundaries)
buckets_min = [np.iinfo(np.int32).min] 
              + boundaries
\end{lstlisting}
\fi

\noindent\textbf{Incorrect Memory Allocation:} A MXNet user reported that they were unable to use multiple GPUs for model training, while a single GPU worked\footnote{\scriptsize{\url{https://github.com/apache/incubator-mxnet/issues/7000}}}. A contributor discovered that the issue was due to an inability to use pinned memory for those GPUs. The patch counts the number of GPUs and ensures that their pinned memory is used during training.
%\textbf{Memory Allocation (Incubator-Mxnet)}: In Incubator-Mxnet GitHub repository, a user reported that the training process in multi-GPU mode could not be complete. But the user`s program is working on the single GPU successfully. After debugging the code, the user realized the reason for this problem is that the pinned memory of GPU cannot use during the training process. The Incubator-Mxnet system already notices this issue and released a patch to fix the error about the pinned memory never being used. But this patch cannot fix the user`s issue. After discussed with the user, the developer of the system realized that this issue could not be covered well by the old patch. And the developer released a new patch to fix this issue: \ref{appendix:a}\\
%In this patch, the developer focus on the problem about code not using pinned memory. The developer adds ‘CUDA\_CALL’ and ‘CHECK\_GT’ functions in the storage file to count the number of GPU device and make sure that one or more GPUs` pinned memory could be used during the process. \\

These examples illustrate how system-related issues affect ML frameworks, illustrating how different hardware configurations, memory and resource constraints, and limited testing of APIs can hinder the use of ML-based systems. 

%From these , we learned some critical factors affect the real reason for system issues from the issues' information and the patch of these issues. Some system issues happened after the users applied a new hardware or update the version of the software platform, and these issues never happened before the users applied these new functions to their program. This kind of problems point out the update delay of the software 2.0 and lead to the system lack of support for new functions. 
\iffalse
\vspace{-10pt}
\begin{lstlisting}[language=diff]
#if MXNET_USE_CUDA
+ CUDA_CALL(cudaGetDeviceCount
+           (&num_gpu_device));
+ CHECK_GT(num_gpu_device, 0) << 
+       "GPU usage requires at least 1 GPU";
 ptr = new storage::GPUPooledStorageManager();
#else
 LOG(FATAL) << "Compile with USE_CUDA=1 to 
                 enable GPU usage";
\end{lstlisting}
\fi

\begin{table}[!t]
\caption{The selected frameworks and indicators.}
\label{table:system1}
\vspace{-10pt}
\begin{center}
\resizebox{\columnwidth}{!}{%
\begin{tabular}{llllllll}
\toprule
\textbf{Framework} & \textbf{Domain} & \textbf{Watches} & \textbf{Stars} & \textbf{Forks} & \textbf{Commits} & \textbf{Contributors} & \textbf{Issues}\\
\midrule
TensorFlow & Machine Learning  & 8585 & 127514 & 74602 & 55724 & 1987 & 28 \\
Torch7 & Scientific Computing  & 664 & 8292 & 2331 & 1337 & 131 & 10\\
Caffe2 & Deep Learning & 559 & 8446 & 2130 & 3680 & 194 & 7\\
PyTorch & Machine Learning & 1244 & 27999 & 6667 & 17915 & 1039 & 11\\
Theano & Scientific Computing & 590 & 8786 & 2483 & 28080 & 332 & 10\\
OpenCV &  Computer Vision  & 2444 & 34673 & 25213 & 26492 & 999 & 12\\
Keras & Deep Learning & 2013 & 41115 & 15652 & 5110 & 794 & 10\\
Chainer & Deep Learning & 321 & 4778 & 1263 & 26356 & 227 & 10\\
CNTK & Deep Learning & 1386 & 16110 & 4267 & 16090 & 199 & 10\\
MxNet & Deep Learning & 1173 & 16856 & 6017 & 9585 & 690 & 13\\
 \hline
React & UI & 6632 & 129558 & 23817 & 10955 & 1295 & 40\\
ETCD & Database & 1268 & 24937 & 5051 & 15102 & 495 & 25\\
Flutter & Mobile & 2269 & 64674 & 7241 & 14264 & 393 & 60\\
Rancher & Container & 602 & 11549 & 1275 & 2686 & 57 & 70\\
iPython & Notebook & 832 & 13576 & 3812 & 23811 & 592 & 24\\
Babel & Compiler & 858 & 33145 & 3511 & 12405 & 726 & 33\\
AWS-CLI & Cloud & 564 & 8029 & 1718 & 6963 & 197 & 11\\
Drone & DevOps & 585 & 18344 & 1800 & 3436 & 241 & 9\\
OSQuery & Operating System & 707 & 14187 & 1721 & 5005 & 264 & 10\\
Grafana & Log System & 1212 & 28857 & 5410 & 21934 & 868 & 50\\
\bottomrule
\end{tabular}}
\end{center}
\vspace{-10pt}
\end{table}

\vspace{-0.1in}
\section{Methodology for Issue Sampling}\label{sec:method}

%\greg{I swapped the order of Methodology and Hypotheses. I thought it made more sense. Please reverse the order again if you think it would be clearer the other way}

To study differences in system-related issues between ML and traditional frameworks, we sample frameworks and their issues, classify them, and collect additional data.

% This section outlines the methodology used to select systems, collect issue data, and perform classification. 

\subsection{System Selection}

In this study, we target frameworks---rather than individual systems---because the functionality offered by frameworks will be utilized by many systems, and will subsequently impact the behavior of such systems. Further, individual systems typically are developed by a smaller team of developers, have a smaller community of users, and will have fewer reported issues.

We selected ten open-source ML frameworks and ten open-source traditional software frameworks, as shown in Table~\ref{table:system1}. 
% We use various indicators to select the systems \todo{used how? we selected large and popular systems?}, including: the number of watches, stars, forks, contributors, commits, and closed issues.
% These frameworks facilitate building systems by providing libraries of utilities. For example, TensorFlow provides a comprehensive, flexible ecosystem of tools, libraries and community resources for building and deploying ML applications.  
We selected ML frameworks based on their popularity and maturity. The popularity of each system in GitHub can be assessed from the number of stars of a repository~\cite{borges2016predicting}. We sought ML frameworks with a reasonable level of maturity. The selected frameworks have 1k-55k commits and more than a thousand forks. 
For contrast, we selected a matching set of 10 traditional (non-ML) frameworks that (a) come from a variety of different fields, and (b) are reasonably matched to the ML frameworks in terms of their activity and popularity. We devised a series of categories, and chose the most popular systems in each field and collect the indicators listed above for each system's repository. We normalized the collected values for each indicator. Then, we compared indicators for each traditional software framework with those for the ML frameworks. In line with propensity score matching~\cite{rosenbaum1983central}, we choose one software framework from each category that was the most similar to one of the ML frameworks. For instance, React and TensorFlow are considered reasonably similar, as judged by the collected indicators.
Table~\ref{table:system1} lists all frameworks, values for the indicators used in pairing, and the number of issues sampled.
% \todo{cite propensity matching as a method?}

\subsection{Sampling Issues}

To understand the nature of system-related issues in ML frameworks, we need to collect enough data to investigate different types of issues. As we want to ensure sufficient information on each issue and how it was fixed, we focused on closed issues---those already fixed. Studying all system-related issues would be prohibitively costly, so instead we sampled issues from all studied systems. 

% The sample size for each system is shown in Table~\ref{table:system1}.

%After we calculate the sample size (see Table~\ref{table:system1}) for each system (based on the method described in Section~\ref{sec:sample-size}), 
In order to avoid selection bias, we \emph{randomly} sample issues for each system to generate the sample set. We created a Python program based on the REST API provided by GitHub to randomly collect closed issues from each system's repository to generate the data set. In this data set, the data for each issue includes the issue title, the issue description, issue timeline, number of participants in the issue discussion and the discussion of the corresponding pull request, and the number of comments on the issue and the corresponding pull request. We also collect information on the patch that fixes the issue, including the number of code lines changed and the number of files changed. To assess community behavior, we collect the number of participants for each issue that are members of the development team, the members' number of contributions to the project, and the number of contributions made by the creator of the issue-closing pull request. 

After that, to avoid meaningless issues, we set inclusion and exclusion criteria to filter the data set. Issues are \textbf{excluded} if \emph{there are less than three comments}, \emph{a patch is not included}, \emph{the issue status is open}, \emph{the item is not an issue (i.e., a pull request in the issue list, a question, or enhancement suggestion)}, or \emph{the issues has been closed due to lack of activity}. The issue is \textbf{included} if it is a system-related issue (not a functional issue), the patch is valid, and if the description has enough information to classify the issue type. 

% After we collect enough system-related issues, we categorize the selected issues into different categories based on the root cause of the issue, issue-related hardware components, and the impact of the issue on the system. We categorize each issue into 15 different categories, as defined in Table~\ref{table:category1}.  When the number of collected system-related issues is equal to the sample size we calculated for each system, we stop. 

\subsection{Determining Sample Size}\label{sec:sample-size}

We used power statistics to compute an appropriate sample size across all ML frameworks and all traditional software frameworks.
At confidence level of 95 percent, we set the margin of error---how much we can expect our analysis result to reflect the view of the overall population---to 10\% for ML frameworks and 5\% for traditional software frameworks (because the traditional frameworks represent a diverse set of domains, and we, therefore, need more observations to understand them). We use the following formula to calculate the sample size~\cite{kadam2010sample}:
\begin{equation}
\frac{\frac{z^{2}*p(1-p)}{e^{2}}}{1+\frac{z^{2}*p(1-p)}{e^{2}*N}},  
\end{equation}
where $N$ is the population size, $e$ is the margin of error and $z$ is the z-score---the number of standard deviations a given proportion is away from the mean, resulting in 121 samples for ML frameworks and 332 samples for traditional software frameworks.
%The z-score depends on the confidence level. Because we set the confidence level to 95 percent for both groups, the z-score is 1.96.
Finally, we allocate the total sample size to each system in the group based on the percentage of the population of closed issues that belongs to each system. The formula for each system's sample size is $SS= GS*(\frac{SI}{GI})$, where $SS$ is the system's sample size, $GS$ is equal to the group sample size, $SI$ represents the total number of closed issues for the system, and $GI$ is the total number of closed issues for the group, resulting in the sample sizes listed in Table~\ref{table:system1}. 

\section{Exploratory Hypotheses}\label{sec:theory}

Our research design is exploratory, but we guide our research using research questions and hypotheses (conjectures) shaped by personal experience in developing large-scale ML systems, interacting with ML developers in industry, and a literature and open source issue review of how ML systems differ from traditional software systems. We will explain our expectations and use them to guide our analysis.

% \todo{this is pretty hard to square: why do we have hypotheses in an exploratory design? where do the hypotheses come from (are they grounded in any meaningful way like interviews or literature survey) and why don't we just to hypothesis testing to test these hypotheses, rather than to explore data? typically the outcome of exploratory research would be hypotheses that can be subsequently tested.}

% In this section, we introduce the research questions that our study aims to answer. For each, we outline the hypotheses that we pose to guide our exploration of the differences in how system-related issues affect ML and traditional software frameworks.

\definecolor{shadecolor}{rgb}{0.9,0.9,0.9} 
\begin{shaded}
\noindent\textbf{RQ1: What differences can be seen in the types and distribution of issues in ML versus traditional frameworks?}
\end{shaded}

This research question allows us to better understand whether particular types of issues are unique to ML frameworks, or differ in frequency of occurrence. This helps us understand whether system-related issues affect ML frameworks differently than traditional software frameworks, and what types of issues developers can expect to see. This allows better risk planning and allocation of resources. In this question, we examine a hypothesis about the distribution of issues.

% To answer RQ1, we focus on two hypotheses that investigates the distribution of system-related issues' categories in Software 2.0 and 1.0 (H1).

%In this research question, we focus on the selected system related issues categories and cross-compare the categories between the software 2.0 and software 1.0. We categorized the systems related issues. Each category represents a specific type of issue. The issue category reflects the root cause of an issue. The number of issues in each category for different systems could help us learn what types of issues are important, allowing us to understand if software 2.0 categorically differ from software 1.0. This would allow us to provide more instructive information to the developers and users of software 2.0 systems when they are facing new system related issues. To answer RQ1, we analyze two research hypotheses that investigates the distribution of system-related issues' categories in the same group (H1), and the difference of categories distribution between software 2.0 and 1.0 (H2).

%\greg{I don't really understand the difference between H1 and H2. The text of these two hypotheses say the same thing. I think the difference you are tying to get across is that H1 is within-group (ML systems vs other ML systems, trad systems vs trad systems) and H2 is across-groups (ML systems vs trad systems). However, your text here does not get this across and your analysis later on for both hypotheses compares the two groups as well. I suggest you have a single hypothesis here - H1. The results later make no meaningful distinction.}

\noindent\textbf{H1: There are categories of system-related issues that occur more frequently or uniquely in ML frameworks, and categories that occur more frequently or uniquely in traditional software frameworks.}

ML-based systems differ in many aspects from traditional software, and proper execution relies on choosing a model, training it, tuning parameters, and correctly executing prediction processes. We suspect that ML frameworks will suffer from system-related issues that occur rarely, if at all, in traditional software. Likewise, certain issues in traditional frameworks may occur rarely or be irrelevant to ML frameworks.

% To evaluate this hypothesis, we categorized all selected system-related issues for each system, and compared the result between the two paradigms to characterize differences in the distribution of the occurrence of issues. 

\definecolor{shadecolor}{rgb}{0.9,0.9,0.9} 
\begin{shaded}
\noindent\textbf{RQ2: Are system-related issues more difficult to fix in ML frameworks than in traditional frameworks?}
\end{shaded}

In some ways, the development of ML frameworks is more complex and less mature than traditional software frameworks. This, in turn, may affect the difficulty of addressing system-related issues. In this question, we examine two hypotheses about the difficulty of issue correction and reproduction. 

\noindent\textbf{H2: There are categories of system-related issues in ML frameworks that are more difficult to fix than in traditional software frameworks.}

ML frameworks have a large volume of input data, complicated algorithms, are built on complex models, and require configuration. These characteristics may impact the difficulty of fixing issues. We also wish to understand whether differences in difficulty are categorical---ML vs traditional frameworks---or dependent on framework-specific factors. For example, TensorFlow supports multiple GPUs, while Theano is bound to a single GPU by default. 

%are the reason why we hypothesize that there are categories of issues that more difficult to get fixed in Software 2.0. We want to examine whether this is true, and if it is, understand why certain types of issues are more difficult to address. The similar or different difficulty level for a specific category between the software 2.0 and software 1.0 or between the different system in the same group could reflect the system characteristics in the real-world applications. For example, TensorFlow could support computational distribution in multiple GPU or CPU, but Theano only supports single GPU in CUDA and cuDNN. This system level difference leads to the difficulty level of deployment-related issues might be different in these two systems. 

% To evaluate this hypothesis, we measure difficulty of fixing an issue from two aspects---human effort and the size of the issue-fixing patch. Human effort can be measured in terms of the time needed to close the issue, and the number of comments in the issue report and corresponding pull requests. Patch change magnitude is measured in terms of the number of code lines and files changed in the patch.

\noindent\textbf{H3: System-related issues are easier to reproduce in traditional frameworks than in ML frameworks. Issue reports in ML frameworks require that the reporter offer additional information in order to reproduce and debug the issue.}

Issues in ML frameworks may arise from a more diverse pool of configurations, hardware platforms, and deployment environments than in traditional frameworks. To reproduce and debug issues, developers may require additional information from the reporter. 

% Compare with the Software 1.0, the system-related issues involved in more complexity deployment environment, unmanageable configuration set, etc. are more common in Software 2.0. The developers need to get all specific information to make sure they can reproduce the issue in the same situation. But the issues in software 1.0 more related to the user’s application in the existing framework, the logic and function in users code is more important than the hardware version.

% To evaluate this hypothesis, we collect four different indicators of the quantity of information in the report, including the total words of the issue's description, the number of comments before developers can reproduce the issue, the number of code lines attached in the issue report, and the number of attached files (i.e., log files). We compare these four indicators between the two groups of systems.

\definecolor{shadecolor}{rgb}{0.9,0.9,0.9} 
\begin{shaded}
\noindent\textbf{RQ3: Are there differences in how communities behave when identifying and fixing system-related issues between ML frameworks and traditional frameworks?}
\end{shaded}

We investigated the behavior of the open-source communities building the studied frameworks. To answer RQ3, we investigate three hypotheses about community behavior, examining participant specialization and experience (H4), discussion activity (H5), and the impact of activity level on the issue-fixing process (H6).

\noindent\textbf{H4: The participants in issue discussion in ML frameworks are more experienced, are more specialized in their knowledge, and attract more popularity than participants in discussions in traditional frameworks.}

As ML frameworks incorporate complex algorithms, the developers of such systems require appropriate specialization in their expertise. Solving system-related issues requires a deep understanding of the complex underlying algorithms (e.g., distributed training). This means that active developers of such systems may be more senior than developers of generic systems and may have certain specific areas of expertise. Because ML represents a new paradigm, users may---in turn---pay more attention to the developers of the systems and their contributions to ML frameworks.

%Software 2.0 as the new type of systems, the users might pay more attention to the system-related issues, and there has more users concern the developer's solution. On the other hand, the developers have a deeper understanding of these issues and have more related experience in this area.

% To evaluate this hypothesis, we collected factors related to the participants in the issue discussion and the particular developer who issues the issue-closing pull request. We track the number of followers that the pull request creator has in GitHub, the number of years that participants have owned their GitHub account, and the role of the participants in the real world. We then compare these three aspects between the two groups of systems.

\noindent\textbf{H5: Discussion of system-related issues attracts a greater number of non-developer users in ML frameworks.}

As ML frameworks are currently attracting a lot of attention, there exists the possibility that issues discussion also attracts a higher level of participation from users who are not part of the development team. Issues may affect a greater number of users, who in turn may experience the same or similar issues in a greater variety of contexts. 
% As a result, we hypothesize that more users will take part in issue discussion in Software 2.0. 

%Software 2.0 has become more and more popular in a variety of industries, and the users could deliver their applications which from diverse area base on the Software 2.0. The different users deal with a similar issue are more often in Software 2.0, since the system-related issues are more specific base on the issue reporter's application in software 1.0. In this way, the Software 2.0 users could provide the experience for the similar issues they deal with before in a different environment or other related useful ideas. As a consequence, there has more unique users who provided comments on the issues' report in software 2.0.

% To evaluate this hypothesis, we collect the number of unique users who provide comments on the issue list. We exclude the developers from this count. We then compare the number of participants between the two paradigms by system.

\noindent\textbf{H6: ML frameworks require a more active developer community than traditional software to fix system-related issues.}

% \greg{The description and motivation for this hypothesis were still confusing. I've tried to revise this based on *my* understanding. Make sure this matches}

The complexity of underlying ML algorithms, increase in need for specialized knowledge, and variety of deployment environments for ML may, in turn, require a more active community of developers in order to address system-related issues. More developers may need to take part in discussion, contribute to the project, and make pull requests in order to maintain a healthy, functioning system.

%The system-related issues in Software 2.0 related to different hardware components, flexible deployment environment, and complex model/ algorithm. It makes the developer in Software 2.0 need more specific experience and knowledge to fix the system-related issues. So, the developers with different experience contribute more to the various categories issues. We hypothesize that the developer activity level of a system in Software 2.0 is higher than the Software 1.0. 

% We measure activity level in three ways---the percentage of members of the development team that take part in issue discussion, the number of contributions a member of the development team makes in the year an issue was discussed, and the number of contributions that pull request creators make in the year an issue was discussed. We compare activity level for Software 2.0 systems with Software 1.0 systems. 

\section{RQ1 --- Issue Characterization}\label{sec:rq1}

\begin{table}[!t]
\caption{Percentage and number of issues in each category. Bolded cells (in all tables) show significantly differing distribution between groups (P-Value $<$ 0.05, One-Way ANOVA).}
\label{table:category2}
\vspace{-10pt}
\begin{center}
\begin{scriptsize}
\begin{tabular}{lll}
\toprule
\textbf{Category} & \textbf{ML} & \textbf{Traditional}\\
\midrule
API Mismatch (API) & 13\% (16) & 15\% (56)\\
Configuration Error (Config) & \textbf{2\% (2)} & \textbf{41\% (148)}\\
Compilation Error (Compl) & 2\% (2) & 0\% (0)\\
Connection Error (Conn) & 0\% (0) & 1\% (4)\\
Data Race (Race) & 1\% (1) & 0\% (0)\\
Execution Error (Exec) & 1\% (1) & 0\% (0)\\
Hardware-Architecture Mismatch (HA) & 1\% (1) & 0\% (0)\\
Incorrect Memory Allocation (MA)  & \textbf{5\% (6)} & \textbf{2\% (8)}\\
I/O Slowdown (I/O)  & 9\% (11) & 5\% (17)\\
Memory Leak (ML) & \textbf{30\% (36)} & \textbf{14\% (50)}\\
Model/Data Conversion (Conv) & 1\% (1) & 0\% (0)\\
Multi-Threading Error (MT) & \textbf{13\% (16)} & \textbf{4\% (13)}\\
Performance Regression (PR) & \textbf{20\% (24)} & \textbf{12\% (42)}\\
Slow Synchronization (SYNC) & 3\% (4) & 7\% (24)\\
\bottomrule
\end{tabular}
\end{scriptsize}
\end{center}
\end{table}

\begin{figure}[!t]
   \centering
   \includegraphics[width=0.6\linewidth]{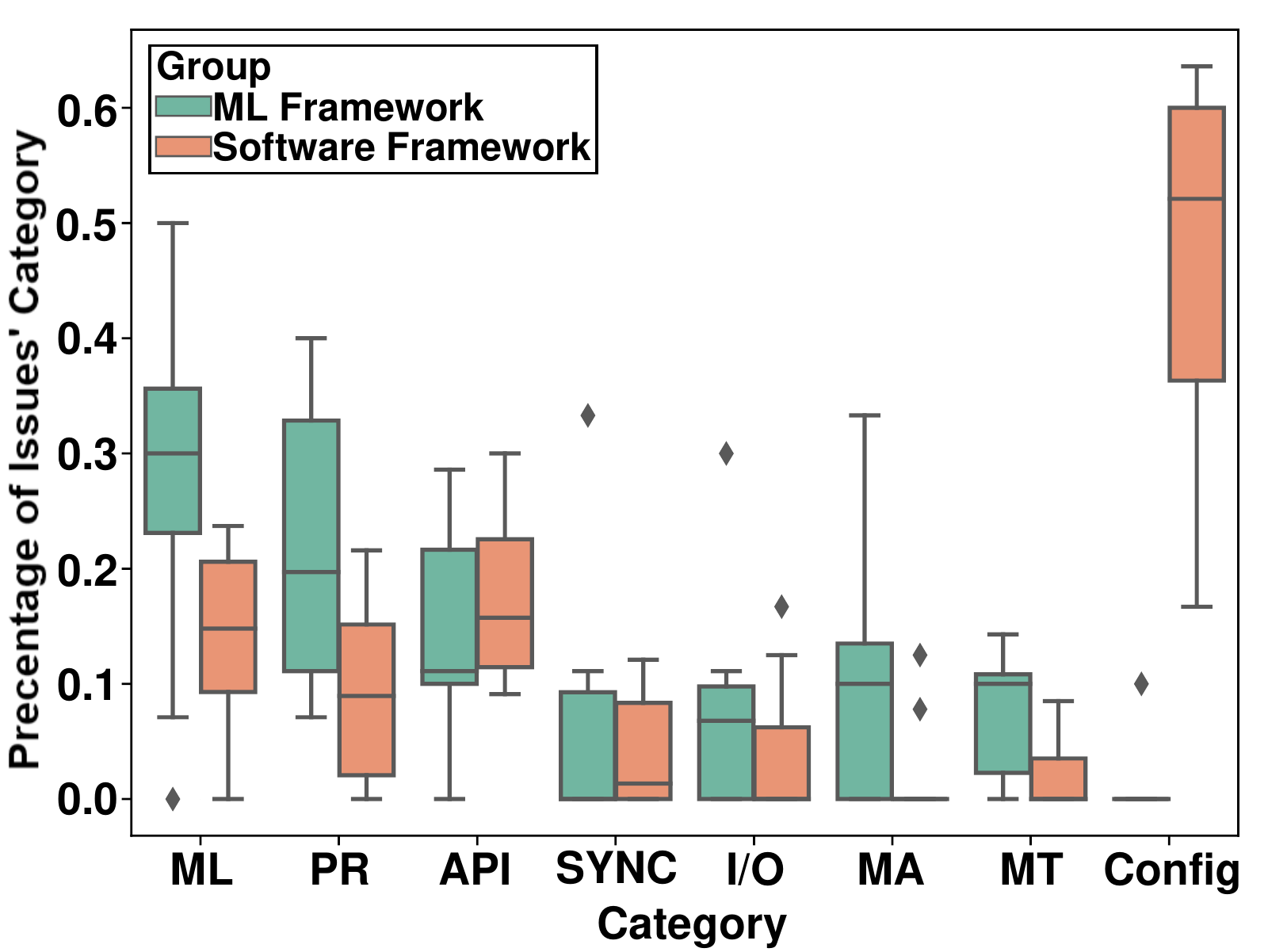} 
   \vspace{-15pt}
   \caption{Percentage of issues belonging to categories.}
   \label{fig:rq1}
\end{figure}

Our first research questions asks about the differences in the types and distribution of issues between ML and traditional software frameworks. More specifically, hypothesis 1 speculates whether the exist statistically significant differences in types and distribution of issues. Table~\ref{table:category2} show the percentage of the total issue pool and raw number of issues for each category for the two paradigms. Figure~\ref{fig:rq1} shows boxplots of the percentage of issues belonging to each category for each system in the two paradigms. For clarity, only common categories are shown. 

From Figure~\ref{fig:rq1}, we can see that there are a number of issue types that occur quite often in both paradigms---including memory leaks, performance regression, API mismatch, slow synchronization, and multi-threading errors. However, there are categories for which there are obvious differences in frequency. Configuration errors are quite common for traditional frameworks, making up 41\% of the total issue pool, with a median of close to 50\% on a per-system basis. They are vanishingly rare for ML frameworks, only making up 2\% of the total pool. On a per-system basis, incorrect memory allocation makes up a median of 10\% of the issue pool for ML frameworks, and is rarer for traditional frameworks. 

To more clearly understand the major areas of difference between the system paradigms, we used a one-way analysis of variance (ANOVA) to compare the distributions of fault types between groups of systems. In Table~\ref{table:category2}, we bold the categories where significant difference was shown between the groups with p-value $<$ 0.05. We find that configuration errors occur significantly more often in traditional frameworks. Systems of both categories are ``configured'', in the sense that their execution depends on can vary depending on certain adjustable factors. However, in traditional software, configuration tends to be explicit, based on providing values in a file or through the command line. For example, a user of AWS-CLI reported an issue that occurs when a space character appears in a provided profile name\footnote{\scriptsize{\url{https://github.com/aws/aws-cli/issues/2806}}}. Such issues are more rare in ML frameworks, where a user rarely directly adjusts values in a file. In ML frameworks, ``configuration'' tends to be more implicit, where---for example---behavior varies based on a chosen hardware platform or training data. This leads to other issues, as we will discuss, but reduced the potential for explicit configuration issues.

We also find that incorrect memory allocation, memory leaks, multi-threading errors, and performance regressions occur significantly more often in ML frameworks. ML systems must process, manage, and make decisions using massive sets of data. Such algorithms must be multi-threaded, in order to rapidly process subsets of the dataset in parallel~\cite{Dean08:MSD}. Likewise, the volume of data and need to store and access it efficiently requires careful management of memory. As a result, threading and memory errors will likely occur more often. Our observations bear this out. The field of ML evolves rapidly, and the popularity of such systems has led to an ever-expanding userbase. The need for rapid evolution may also explain the increased frequency of performance regressions.

In our random sample, there were several types of issues uniquely observed in ML---as can be seen in Table~\ref{table:category2}. These include data races, execution errors, hardware-architecture mismatch, model/data conversion, and unexpected resource usage. None of these types were common, and most of these can---without doubt---occur in traditional frameworks as well. However, several of these issue categories closely relate to important facets of ML systems, and may occur more commonly as a result. For example, 
hardware-architecture mismatch can occur because of the variety of hardware configurations being used in ML. Many ML platforms can use GPUs for efficient data processing, particularly using NVIDIA's CUDA platform~\cite{Sanders10:CEI}. As an example, an issue encountered in this study, from OpenCV, occurred because the system did not support the version of CUDA used by the GPU in the user's configuration\footnote{\scriptsize{\url{https://github.com/opencv/opencv/issues/7375}}}.

\definecolor{shadecolor}{rgb}{0.9,0.9,0.9} 
\begin{shaded}
%\textbf{Summary}: Configuration error issue is the most significant difference in category distribution between the software 2.0 and 1.0. Memory leak issue, API issue, performance regression issue, SYNC issue, I/O issue, memory allocation error issue, and multi-thread issue as the common categories for both group, the developers or users from these two groups all need to take care of these categories. 
\textbf{Summary}: Incorrect memory allocation, memory leaks, multi-threading errors, and performance regression occur more commonly in ML frameworks---likely due to the need to manage large quantities of data in memory and the rapid pace of system enhancement. Increased dependence on hardware selection, like the GPU, can also lead to issues. Configuration errors are very common in traditional frameworks, but rarely occur in ML, as such frameworks tend to offer fewer explicit user-defined configuration options.
\end{shaded}

\section{RQ2 --- Issue Difficulty}

Our second research question asks whether system-related issues are more difficult to fix in ML frameworks than in traditional software frameworks. We focus on the six categories of issues with a reasonable number of samples for both traditional software and ML frameworks: memory leaks, performance regressions, API mismatch, I/O slowdown, incorrect memory allocation, synchronization, and multi-threading errors. We guide our analysis using two exploratory hypotheses: (H2) that there are categories of issues that are more difficult to fix in ML, and (H3), that issue reporters must provide additional information to developers of ML frameworks. 

\subsection{H2---Issues are More Difficult to Fix}

\begin{table*}[!t]
\caption{Median values for each collected data regarding issue difficulty for Hypothesis 2.} % Bolded cells under an indicator show significantly differing distribution between groups (P-Value $<$ 0.05, One-Way ANOVA).}
\label{table:hypothesis2}
\vspace{-10pt}
\begin{center}
\begin{scriptsize}
\begin{tabular}{lllllllllllllll}
\toprule
& \multicolumn{2}{c}{\textbf{Days}} & \multicolumn{2}{c}{\textbf{Comments-Rep}} & \multicolumn{2}{c}{\textbf{Comments---PR}} & \multicolumn{2}{c}{\textbf{Participants---Rep}} & \multicolumn{2}{c}{\textbf{Participants---PR}} & \multicolumn{2}{c}{\textbf{LOC}} & \multicolumn{2}{c}{\textbf{Files}} \\ 
\textbf{Category} & \textbf{ML} & \textbf{Trad} & \textbf{ML} & \textbf{Trad} & \textbf{ML} & \textbf{Trad} & \textbf{ML} & \textbf{Trad} & \textbf{ML} & \textbf{Trad} & \textbf{ML} & \textbf{Trad} & \textbf{ML} & \textbf{Trad}\\
\midrule
API Mismatch (API) & \textbf{47.0} & \textbf{7.5} & \textbf{9.0} & \textbf{7.0} & 1.0 & 3.0 & \textbf{4.0} & \textbf{3.0} & 2.0 & 2.5 & 24.0 & 36.0 & 2.0 & 2.0 \\
Incorrect Memory Allocation (MA) & 8.0 & 11.5 & 6.0 & 6.5 & 1.5 & 0.0 & 3.0 & 2.5 & \textbf{3.0} & \textbf{0.0} & 8.0 & 23.5 & 1.5 & 2.5 \\
I/O Slowdown (I/O) & 13.0 & 7.0 & 4.0 & 7.0 & 1.0 & 5.0 & 3.5 & 3.0 & 3.5 & 3.0 & 30.5 & 58.0 & 1.0 & 2.0 \\
Memory Leak (ML) & 10.0 & 10.0 & 6.0 & 6.0 & 0.0 & 6.0 & 5.0 & 3.0 & \textbf{2.0} & \textbf{4.0} & 28.0 & 34.0 & 2.0 & 3.0 \\
Multi-Threading Error (MT) & 32.0 & 10.0 & 9.0 & 9.0 & 6.0 & 1.0 & 3.0 & 2.0 & 3.0 & 1.0 & 45.0 & 16.0 & 2.0 & 1.0 \\
Performance Regression (PR) & 12.5 & 12.0 & 9.0 & 8.0 & 6.0 & 3.0 & 4.5 & 4.0 & 5.0 & 2.0 & 25.5 & 37.5  & 2.0 & 2.0 \\
\textbf{Overall} & 11.0 & 8.0 & 7.0 & 7.0 & 1.0 & 3.0 & 4.0 & 3.0 & 3.0 & 3.0 & 26.0 & 23.0 & 2.0 & 2.0 \\
\bottomrule
\end{tabular}
\end{scriptsize}
\end{center}
\end{table*}

\begin{figure}[!t]
   \centering
   \includegraphics[width=0.6\linewidth]{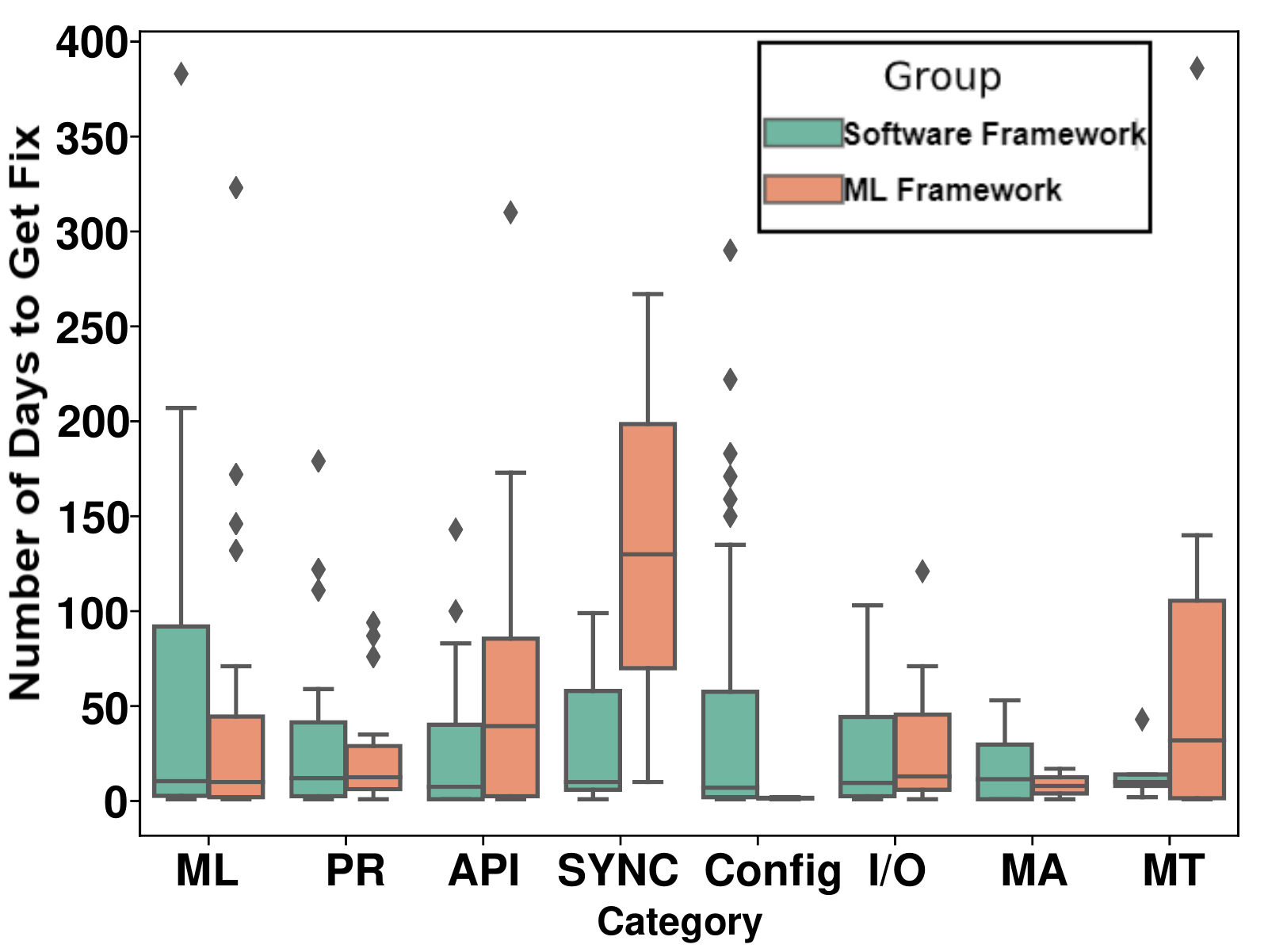}\vspace{-15pt}
   \caption{Number of days from issue creation to close.}
   \label{fig:days}
\end{figure}

\begin{figure}[!t]
   \centering
   \vspace{-5pt}
   \includegraphics[width=0.6\linewidth]{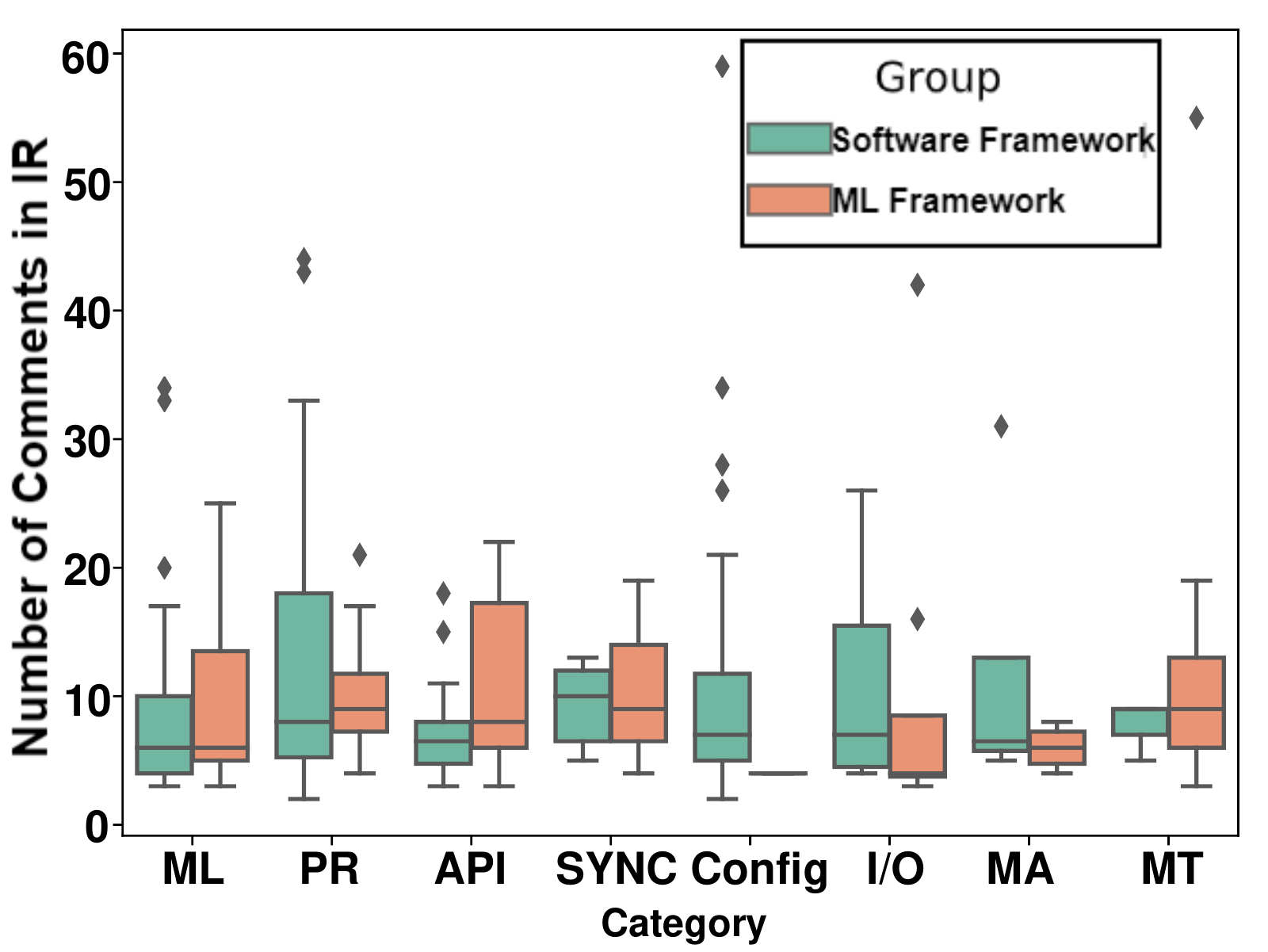}\vspace{-15pt}
   \caption{Number of comments in the issue report.}
   \label{fig:comments}
\end{figure}

\begin{figure}[!t]
   \centering
   \includegraphics[width=0.6\linewidth]{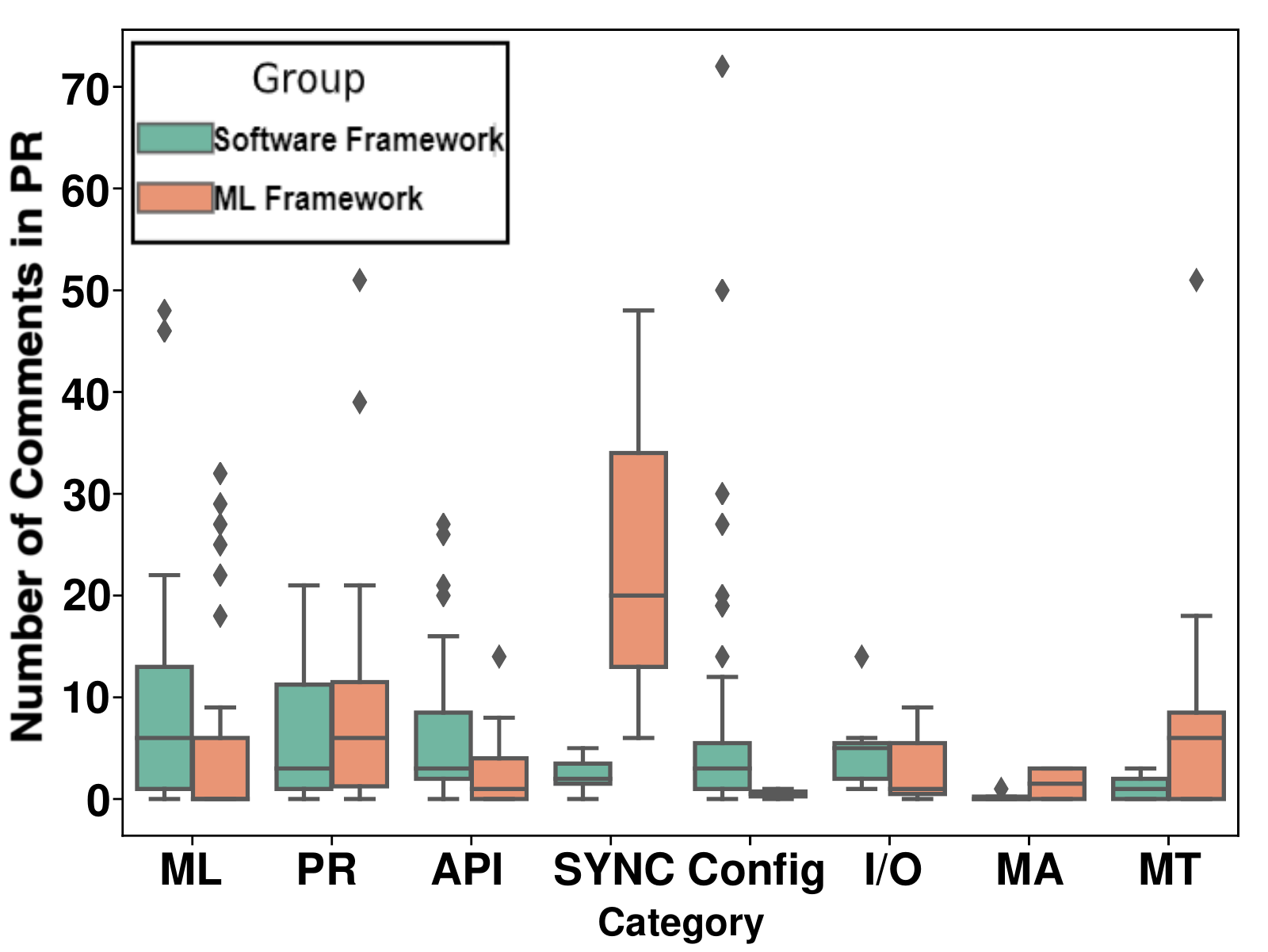}\vspace{-15pt}
   \caption{Number of comments in the pull request.}
   \label{fig:commentsPull}
\end{figure}

Hypothesis 2 speculates that certain categories of issues are more difficult to fix in ML than in software frameworks. We gathered seven indicators that, together, present an approximation of the effort required to fix an issue. These indicators include the number of days between issue creation and closure, the number of comments on the issue report, the number of participants in the issue report discussion, the number of comments on the pull request closing the issue, the number of participants involved in discussion of the pull request, the number of lines of code changes in the patch fixing the issue, and the number of files changed. Table~\ref{table:hypothesis2} lists the median values for each indicator for six issue types and a summary across all types of issues. We provide details about the distribution of these indicators in Figures~\ref{fig:days},~\ref{fig:comments}, and~\ref{fig:commentsPull}.

% , ~\ref{fig:participants}, ~\ref{fig:participantsPull}, ~\ref{fig:lines}, and ~\ref{fig:files}.

% ... we show boxplots for number of days, comments, participants, code line changed in patch, files changed in patch. Plots for other measures are omitted due to space constraints, and because they do not benefit discussion.

From Table~\ref{table:hypothesis2}, we can see that---overall---issues seem to take slightly longer to be fixed in ML frameworks, with a median of 11 days versus 8 days. They also tend to require slightly larger patches (26 LOC versus 23). However, neither of these indicators show a significant difference according to the ANOVA test, and many of the other indicators---comments on the report, participants in the PR, and number of files changed---have the same median. Therefore, there is little we can conclude about issue difficulty overall. \textit{ML framework issues, overall, are not more difficult to fix.} It is, however, worth looking more deeply at individual categories of issues. 
API mismatch issues do take longer to fix in ML, taking a median of 57 days (compared to 7.5 days in traditional frameworks) and demonstrating significant difference in the ANOVA test. Likewise, there tends to be more discussion on the issue report, from more participants. From this, we can speculate that API issues may not actually be more difficult to fix in terms of traditional code changes. Rather, they may be more difficult because the APIs themselves are evolving rapidly following debate in an active, opinionated community. The long median time to fix, and the larger number of comments on issue reports, suggest that API mismatch issues require debate and community deliberation to determine if they are, in fact, actual problems or misuse of the framework. Multiple sampled issues show debate between contributors before consensus is reached on whether there is an issue\footnote{\scriptsize{\url{https://github.com/torch/torch7/issues/281}}} \footnote{\scriptsize{\url{https://github.com/opencv/opencv/issues/6081}}} \footnote{\scriptsize{\url{https://github.com/tensorflow/tensorflow/issues/25882}}}.
Once developers agree that there is a bug, changes to the API---which have the potential to affect a large number of users---require further debate. 

This is also suggested in Figures~\ref{fig:days} and~\ref{fig:comments}, where there is a large variance in ML frameworks for number of days and number of comments. This variance suggests some contention in the discussion of API mismatch issues. By contrast, Figure~\ref{fig:commentsPull} shows less variance for ML than traditional frameworks in terms of the number of comments in the pull request. By the time a pull request is filed, it tends to be rapidly accepted. 

Incorrect memory allocation issues are more common in ML frameworks, but do not necessarily appear to be more difficult to solve. However, pull requests fixing such issues seem to attract some debate, with significantly more participants involved at the pull request level (confirmed by ANOVA). 

%Multi-threading errors do appear to be more somewhat more difficult to fix in ML frameworks, with higher median time to fix, comments on the pull request, participation in the issue and pull request discussion, LOC changed, and files changed. None of these indicators showed significant differences in the ANOVA test, but higher medians and the increased frequency of occurrence suggests that developers do need to plan for the need to correct multi-threading. 

%We observer no significant trends for I/O slowdown issues. Such issues take somewhat longer to fix (median of 13 to seven days), but require less discussion on the issue and smaller patches. Likewise, performance regressions require slightly longer to fix, and more discussion and participation in the issue and pull request, but smaller patches. The medians are slightly higher for ML, but ANOVA tests do not suggest significant differences. 

Memory leaks are fixed in approximately the same amount of time, with the same quantity of issue discussion. In fact, despite happening more frequently in ML frameworks, memory leaks may be slightly \textit{easier} to fix, with significantly fewer participants in the pull request. Given increased frequency of memory leaks, developers may have a more immediate understanding of how to solve such issues using automated tools, and the fixes for such issues may be accepted with little need for community debate.

\definecolor{shadecolor}{rgb}{0.9,0.9,0.9} 
\begin{shaded}
\textbf{Summary:} Broadly, issues in ML do not appear to be significantly more difficult to address. API mismatch issues require significant time and discussion to fix, reflecting rapidly evolving communities debating how to best evolve their systems. Incorrect memory allocation issues also attract significantly more participants in discussing potential fixes. The most contentious issues reflect an evolving field and an active community. Memory leaks attract less participation at the pull request level, indicating an area where issue commonality leads to quick acceptance of solutions.
\end{shaded}

\subsection{H3---Information Quantity}

\begin{table}[!t]
\caption{Median values for Hypothesis 3.} % Bolded cells under an indicator show significantly differing distribution between groups (P-Value $<$ 0.05, One-Way ANOVA).}
\label{table:hypothesis3}
\vspace{-10pt}
\begin{center}
\begin{scriptsize}
\begin{tabular}{lllllllll}
\toprule
& \multicolumn{2}{c}{\textbf{Comments}} & \multicolumn{2}{c}{\textbf{Words}} & \multicolumn{2}{c}{\textbf{Files Attached}} & \multicolumn{2}{c}{\textbf{Code Attached}}  \\ 
\textbf{Category} & \textbf{ML} & \textbf{Trad} & \textbf{ML} & \textbf{Trad} & \textbf{ML} & \textbf{Trad} & \textbf{ML} & \textbf{Trad} \\
\midrule
API & 2 & 1 & 127 & 104.5 & \textbf{0.5} & \textbf{0} & 21 & 14\\
MA & 0.5 & 2 & 167 & 70 & 0.5 & 1 & 0 & 7 \\
I/O  & 0.5 & 1 & 136.5 & 90 & 0 & 0.5 & 4.5 & 1.5 \\
ML & 0 & 1 & \textbf{131} & \textbf{110} & 0 & 0 & 18 & 17 \\
MT & 2 & 2 & 143 & 115 & 2 & 1 & 36 & 1.5 \\
PR & 0 & 1 & 186 & 124 & 1 & 0 & 0 & 12\\
\textbf{Overall} & 2 & 1 & \textbf{113} & \textbf{93.5} & 0 & 0 & 21 & 12.5\\
\bottomrule
\end{tabular}
\end{scriptsize}
\end{center}
\vspace{-10pt}
\end{table}

% \greg{Figures are a bit hard to read. Also, do we need all of them? If we need space, we can probably drop the one on number of words.}

Our third guiding hypothesis states that more information will be required for developers to reproduce reported system-related issues. We hypothesize this for multiple reasons. ML systems are often somewhat stochastic in nature, behavior is often influenced by subtle environmental factors, and understanding an issue may require specialized understanding of the underlying statistical algorithms. Therefore, we suspect that the reporting user may need to provide detailed information on both the issue and their deployment environment. This could, in turn, contribute to the difficulty of correcting an issue. 

We measure several indicators of the information content that a user must provide. These indicators include the number of comments in the discussion thread before the issue is reproduced. This is determined manually for each sampled issue. We also collect the number of words in the issue description, the number of files attached to the issue report, and the number of lines of code attached to the issue report. Table~\ref{table:hypothesis3} lists the median values for each indicator both for the six issue types with a reasonable number of samples for both system categories and over the full pool of issues. %Figures~\ref{fig:reprComment}-\ref{fig:reprCodes} illustrate results for the six studied issue types.

Overall, as shown in Table~\ref{table:hypothesis3}, ML frameworks require a higher median number of comments before issues are reproduced (2 to 1), number of lines of code attached to the report (21 to 12.5), and number of words in the issue report (113 to 93.5). However, of those, only the number of words shows statistical significance---as demonstrated using the ANOVA test. Therefore, an increase in the amount of information that a user has to provide primarily manifests in terms of the number of words in the issue description. \textit{Users of ML systems provide detailed descriptions of issues to the development community.} This does not necessarily suggest that issues are harder to reproduce or solve in ML, but may instead suggest that the users of such systems are knowledgeable, have more development experience, and may be prepared to offer more background on the issue being reported than the average issue reporter in a software system.

Memory leaks, in particular, require a significantly larger number of words in the issue description. Memory leak issues are not necessarily harder to reproduce, but do require that the user provide a detailed account. This may not reflect the \textit{difficulty} of fixing memory leaks, but rather that the increased frequency of memory leaks in ML better prepares users to report such problems. It is possible that the descriptive initial bug reports help ease acceptance of the pull request, as indicated in the previous section.

API mismatch also requires a significantly higher number of attached files with the issue description.
%From Figures~\ref{fig:reprComment},~\ref{fig:reprFiles}, and~\ref{fig:reprCodes}, we also see quite a bit of variance in the indicators. 
This further suggests that API mismatch issues are difficult to address, and can require debate in the development community---for instance, requiring a higher median number of comments before being confirmed as an issue. The variance between systems is low in terms of the number of words in the description, suggesting along with the higher median that users---up front---provide more information on these issue in ML frameworks. 

\iffalse
\begin{figure}[!htb]
   \centering
   \includegraphics[width=0.5\linewidth]{Figure-h3-1.pdf}\vspace{-15pt}
   \caption{Number of comments before developers reproduce.}
   \label{fig:reprComment}\vspace{-10pt}
\end{figure}

\begin{figure}[!htb]
   \centering
   \includegraphics[width=0.5\linewidth]{Figure-h3-3.pdf}\vspace{-15pt}
   \caption{Number of files attached to the issue report.}
   \label{fig:reprFiles}\vspace{-10pt}
\end{figure}

\begin{figure}[!htb]
   \centering
   \includegraphics[width=0.5\linewidth]{Figure-h3-4.pdf}\vspace{-15pt}
   \caption{LOC attached to the issue report by the reporter.}
   \label{fig:reprCodes}
\end{figure}
\fi

The remaining categories offer little in the way of clear trends. While medians may differ in various ways, the differences are not significant according to the one-way ANOVA tests. 

%Incorrect memory allocations appear to be somewhat easier to reproduce in ML, with lower median number of comments, files attached, and LOC attached---with only a higher median number of words in the description. I/O slowdown requires fewer comments, and files attached, but a more detailed description and more code attached.  Performance regressions require more detailed descriptions, and more files attached, but fewer comments and LOC. 

\definecolor{shadecolor}{rgb}{0.9,0.9,0.9} 
\begin{shaded}
\textbf{Summary}: Users of ML frameworks provide more detailed issue descriptions. Issues are not necessarily harder to reproduce. Instead, users may be more knowledgeable, have more development experience, and may be more prepared to offer background on the issue being reported than in traditional frameworks.
\end{shaded}

\section{RQ3 --- Community Behavior }

Our third research question revolves around the behavior of the open-source communities building the studied systems. 
% Are there differences in how communities behave when identifying and fixing system-related issues between Software 2.0 and Software 1.0?
To answer RQ3, we investigate three hypotheses, examining participant specialization and experience (H4), discussion activity (H5), and the impact of the community activity level on the issue-fixing process (H6). Rather than discussing particular issue types in this question, we focus on the differences between systems.

%\greg{Figures are nearly impossible to read. Omit outliers and increase font size.}

%\greg{Figure 11 (real-world role) still refers to N-ML and ML instead of Software 1.0 and 2.0. Fix this}

\subsection{H4---Participant Experience}

% \begin{table}[!t]
% \caption{Median values for each collected data value for Hypotheses 4 (Followers, Years) and 5 (Unique Users). }
% % Bolded cells under an indicator show significantly differing distribution between groups (P-Value $<$ 0.05, One-Way ANOVA).}
% \label{table:hypothesis4}
% \vspace{0.1pt}
% \begin{center}
% \begin{scriptsize}
% \begin{tabular}{llllllll}
% \toprule
% \multicolumn{2}{c}{\textbf{Followers}} & \multicolumn{2}{c}{\textbf{Years}} & \multicolumn{2}{c}{\textbf{Unique Users}} & \multicolumn{2}{c}{\textbf{Role Identified Participants}}  \\ 
% \textbf{2.0} & \textbf{1.0} & \textbf{2.0} & \textbf{1.0} & \textbf{2.0} & \textbf{1.0} & \textbf{2.0} & \textbf{1.0}\\
% \midrule
% \textbf{62} & \textbf{54} & 7 & 8 & 2 & 2 & 84.38\% & 59.51\% \\
% \bottomrule
% \end{tabular}
% \end{scriptsize}
% \end{center}
% \vspace{-10pt}
% \end{table}

\begin{figure}[!t]
   \centering
   \includegraphics[width=0.5\linewidth]{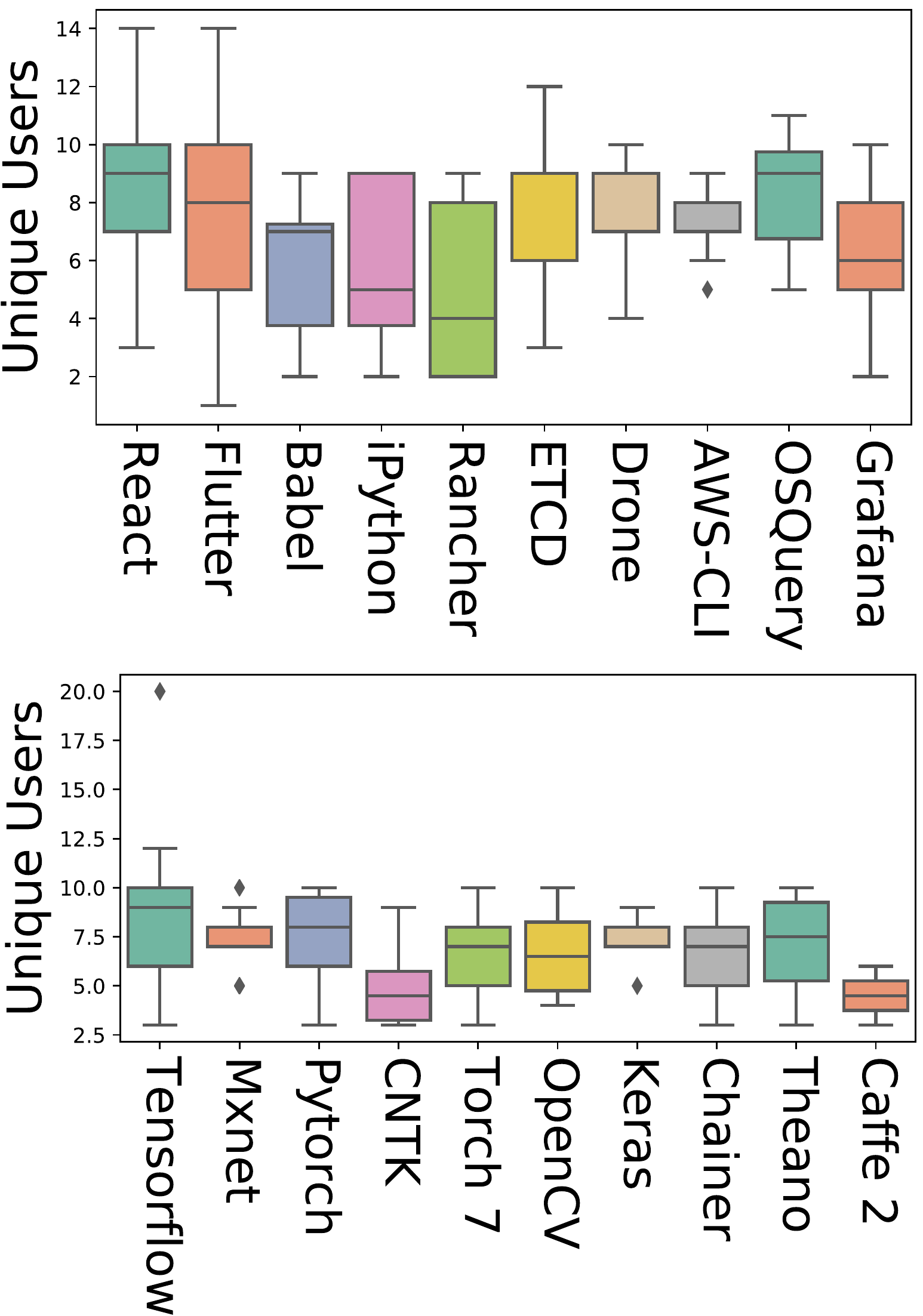} \vspace{-10pt}
   \caption{The participants account history.} 
   \label{fig:Year} 
\end{figure}

\begin{figure}[!t]
   \centering
   \includegraphics[width=0.6\linewidth]{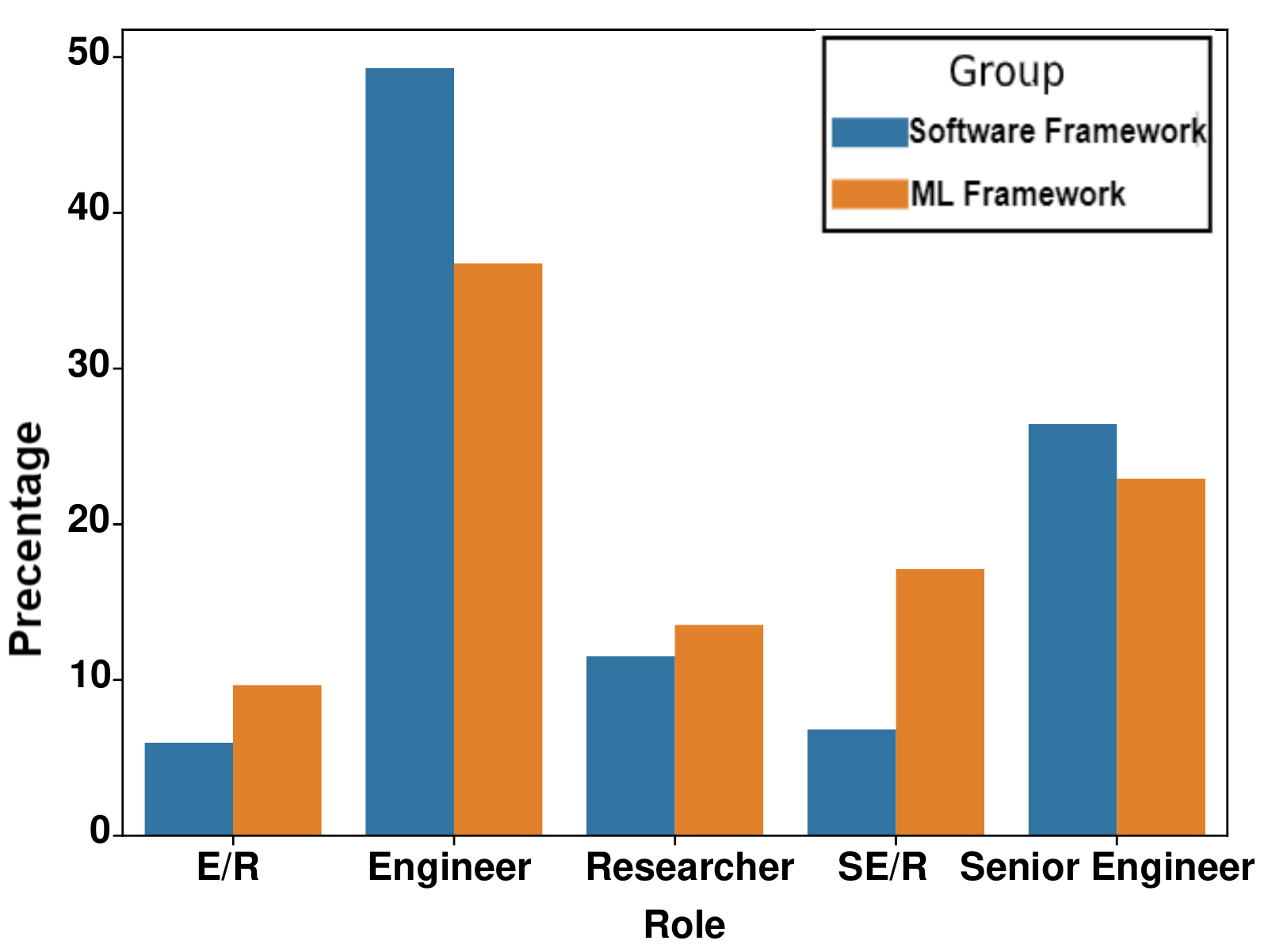} \vspace{-5pt}
   \caption{The real-world role of participants.} 
   \label{fig:Role} \vspace{-10pt}
\end{figure}
   
% \begin{figure}[!t]
% \centering
% \subfigure[Software 2.0]{
% \includegraphics[width=0.7\linewidth]{Figure-h4-followers-ml.pdf}
% }
% \quad
% \subfigure[Software 1.0]{
% \includegraphics[width=0.7\linewidth]{Figure-h4-followers-nml.pdf}
% }
% \caption{Number of followers of participants.}
% \label{fig:Follower} \vspace{-10pt}
% \end{figure}

\iffalse
\begin{figure}[!t]
   \centering
   \includegraphics[width=0.6\linewidth]{Figure-h4-followers-nml-ml.pdf} \vspace{-10pt}
   \caption{Number of followers of participants.} 
   \label{fig:Follower}
\end{figure}\fi
   
Our fourth hypothesis states that the participants in issue discussion in ML frameworks are more experienced and are more specialized in their experience. We measure three indicators for this hypothesis---the number of years that participants in issue discussion have owned their GitHub account, the role of the participants, and the number of followers participants have. %Figures~\ref{fig:Year}-\ref{fig:Follower} illustrate the results for each indicator. 

% , and attract more popularity than participants in Software 1.0 issues

The number of years a user has owned their account partially indicates their development experience. We did not identify any statistically significant results. Our results, shown in Figure~\ref{fig:Year}, indicate quite a bit of variance between systems.
%less variance for ML frameworks, suggesting that---while ML developers may not have more experience than software developers---developers of a particular system tend to have a relatively uniform level of experience. 

% Table~\ref{table:hypothesis4} indicates a slightly higher median for Software 1.0---8 years versus 7---but ANOVA tests do not reveal significant differences.

%The second indicators is the number of years of the participant account history in GitHub. For this indicator, we collect the number of years from the account created to 2019 for each issue participants. This indicator could measure the participants' experience in the GitHub community. Table~\ref{table:hypothesis4} shows that there not exist siginificant difference between the Software 2.0 and 1.0, but the median of participants' account history in the Software 1.0 is slightly higher than Software 2.0. 

%For the detail of the number of years of the participant account history distribution for the systems in each group. Figure~\ref{fig:Year} shows that the system-related issues' participants in Tensorflow, Pytorch, Torch7, OpenCV, Keras, Chainer, and Theano have longer history in the GitHub community than other systems in Software 2.0. For Software 1.0, the system-related issues' participants in React, Flutter, ETCD, Drone, and OSquery have longer history in the GitHub community than other systems in Software 1.0.

Our second indicator revolves around the job role of the participants in issue discussion. For this indicator, we manually gathered the job title for each participant from their GitHub profile or linked personal and LinkedIn pages. Figure~\ref{fig:Role} shows the results of this process. The categories we assigned include Engineer, Researcher, Senior Engineer, and combinations of (Senior) Engineer and Researcher. 
We were able to identify the role of more participants in ML. This potentially suggests that the participants in such projects tend to have a more distinct online identity that ties open-source development into their work role, or where their work role explicitly involves open-source development.  The clearest difference we see between ML and traditional frameworks, as shown in Figure~\ref{fig:Role}, is the increased importance of the role of ``Researcher'' in the participant pool. More participants work as a combination of Engineer and Researcher in ML than in traditional frameworks.

% Because we are able to identify more roles in Software 2.0, Figure~\ref{fig:Role} shows more participants belonging to every category.

%\greg{The Role results need to be better depicted. The most obvious finding of this was that Yang couldn't identify the job role of as many 1.0 participants. That alone could explain the rest of the results, where 2.0 has more items in *every* category. This makes it hard to infer clear results. I would suggest redoing the percentages to be of number that you could identify. For example, Engineer = (\# Engineers) / (\# non-N values). This would make the results much clearer. Then, rewrite this section based on the fixed results.}

%If the participants do not provide this information, we will search their name in Google.com and find their job title. Sometimes there has several different search result. For this situation, we will use their picture, which they used in their GitHub account to match their information. If we cannot identify the participant's information, we use "N" to represent their real-world role. We summarize the participants' job title into five different roles including Engineer, Researcher, Senior Engineer, Engineer/Researcher (E/R), and Senior Engineer/Researcher (SE/R). Figure~\ref{fig:Role} shows the distribution of each participant role for Software 2.0 and 1.0. Compare with Software 1.0, Software 2.0 has significant more Researcher, Engineer & Researcher, Senior Engineer & Researcher provide theirs contributes to the system-related issues. 

Our third indicator is the number of followers that participants in the issue fixing process have on GitHub. This indicates user popularity. We have removed 13 outliers, who had more than 40,000 followers. Overall, ML participants have a higher median number of followers---62 to 54. ANOVA confirms statistical significance. %Software 2.0 developers tend to be more popular on GitHub. 
%As shown in Figure~\ref{fig:Follower}, popularity does vary quite a bit between systems. TensorFlow, Keras, and Caffe 2 show the highest median. 

% CNTK also has a large third quartile, with a relatively low median. The examined Software 1.0 systems have a more restrained level of variance, although React also has a large third quartile. 

\definecolor{shadecolor}{rgb}{0.9,0.9,0.9} 
\begin{shaded}
\textbf{Summary}: Many ML frameworks developers identify as a combination of Engineer and Researcher, while many traditional framework developers identify solely as an Engineer. ML framework developers also tend to be more popular than the developers of traditional frameworks. There is little consistency in how long developers have had GitHub accounts.
\end{shaded}

% ---or omit their job role from their open-source developer identity

%Table~\ref{table:hypothesis4} shows that there not exist a significant difference between Software 2.0 and 1.0 for most of the selected system-related issues. But there have a certain number of participants with significant more followers in the Software 1.0 as the outliers in Figure~\ref{fig:Follower}. For Software 2.0, the system-related issues' participants in Tensorflow, CNTK, and Keras have more followers in the GitHub community than other systems. Beside the Tensorflow, CNTK, and Keras, the number of participants' followers in other systems is at a similar level. Figure~\ref{fig:Follower} shows that the system-related issues' participants in React have more followers in the GitHub community than other systems in Software 1.0. Beside the React, the number of participants' followers in other systems is at a similar level.

%As a result of this hypothesis, the researchers participate in the system-related issues are more often in Software 2.0. But we cannot generate the result like the issues participants are more senior, more experienced in Software 2.0 because there not exist significant difference about the participants' follower and the participants' GitHub accounts history between the Software 2.0 and 1.0.

\begin{figure}[!t]
   \centering
   \includegraphics[width=0.5\linewidth]{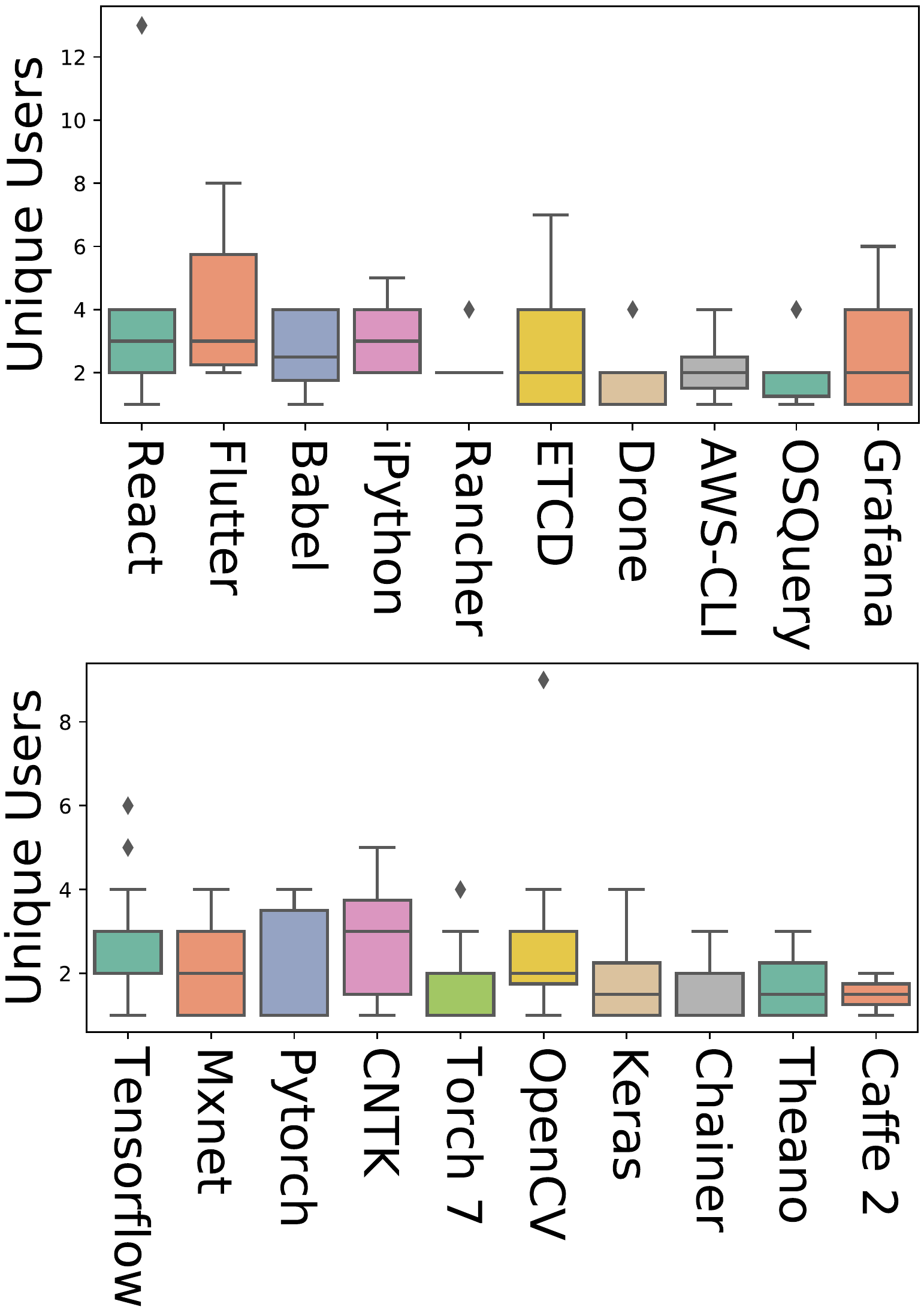} \vspace{-10pt}
   \caption{Number of unique users involved in discussion.} 
   \label{fig:Uusers} \vspace{-10pt}
\end{figure}

\subsection{H5---Non-Developer Users}

% \begin{figure}[!t]
% \centering
% \subfigure[Software 2.0]{
% \includegraphics[width=0.7\linewidth]{Figure-h5-users-ml.pdf}
% }
% \quad
% \subfigure[Software 1.0]{
% \includegraphics[width=0.7\linewidth]{Figure-h5-users-nml.pdf}
% }
% \caption{Number of unique users involved in issue discussion.}
% \label{fig:Uusers}\vspace{-10pt}
% \end{figure}

Our fifth hypothesis states that the discussion of issues attracts a greater number of non-developer users in ML than in software systems. As such systems are currently quite popular, they may have a more active and varied discussion community. To measure the quantity, we collect the number of users from the participant list for each issue. We then omit any users that are listed members of the project development team (who have ``write'' access). An ANOVA test fails to indicate significant differences between the two paradigms. Figure~\ref{fig:Uusers} shows that the number of users that take part in discussion varies quite a bit between frameworks.

% Table~\ref{table:hypothesis4} indicates that Software 2.0 and 1.0 systems have the same median number of non-developer users.
 %For example, 2-3 users tend to take part in TensorFlow discussions---heavily concentrated towards two---while 1-4 take part in CNTK discussions---centered towards three. 

%\greg{Yang - you have the same medians in the table, but your text (commented out below) says that the number is higher in Software 2.0. Did you make a mistake in the table?}
%In this hypothesis, we only focus on the users. Figure~\ref{fig:Uusers} shows the number of unique users as participants in the system-related issues distribution for Software 2.0 and 1.0. Table~\ref{table:hypothesis4} shows that the quantity of unique users in software 1.0 is larger than the Software 2.0.

%For the detail of the number of unique users in the issues' participants distribution for the systems in each group. Figure~\ref{fig:Uusers} shows that the system-related issues in Tensorflow, Pytorch, CNTK, and OpenCV attract more regular users to provide their contribution in the issue fixing process than the other systems in Software 2.0. 

%Figure~\ref{fig:Uusers} shows that the system-related issues in React, Flutter, Babel, and iPython attract more regular users to provide their contribution in the issue fixing process than the other systems in Software 1.0. 

\definecolor{shadecolor}{rgb}{0.9,0.9,0.9} 
\begin{shaded}
\textbf{Summary}: ML issues do not attract significantly more non-developer users to take part in discussion than software frameworks. There is a large amount of variance in the makeup of discussion participants between systems.
\end{shaded}

\subsection{H6---Activity Level}

%\greg{Make sure system names are consistent throughout the paper. Names in some of the figures do NOT match the tables. This is sloppy, and reviewers may get mad at you for it. Make sure you use the actual name from the project page (TensorFlow, not Tensorflow).}

%\greg{Add the number of members to the table. These numbers are presented without full context}

Our final hypothesis is that ML frameworks require a more active developer community than traditional frameworks in order to fix system-related issues. We speculate that more developers may need to take part in discussion, contribute to the project, and make pull requests in order to maintain a healthy, functioning system. 

To measure the activity level, we focus on \textit{members} of each project. The members are people who are part of the organization that owns a project, and that have ``write'' access to the project repository. This list is made available as part of a GitHub project\footnote{\scriptsize{\url{https://github.com/orgs/tensorflow/people}}}. We collect three indicators, including: (1) the percentage of members that take part in issue discussion, (2) the number of contributions made by a member during the year that each issue was reported, and (3) the number of contributions (commits, pull requests) made by the issue's corresponding pull request creator during the year that each pull request was created. 

\begin{table}[!t]
\caption{Community behavior data.}
% The number of members, median percentage of members that take part in issue discussion, median number of contributions in a year from project members, and median number of pull requests in a year from all contributors. }
\label{table:member1}
\vspace{-10pt}
\begin{center}
\resizebox{\columnwidth}{!}{
\begin{tabular}{lllll}
\toprule
\textbf{System} & \# \textbf{Members} & \textbf{\% Participate} & \textbf{\# Contributions} & \textbf{\# PR Contributions}\\
\midrule
TensorFlow & 2 & 33.33\% & 184 & 387\\
Torch7 & 1 & 33.33\% & 2503 & 974\\
Caffe2 & 0 & 0\% & 0 & 448\\
PyTorch & 1 & 25.00\% & 1465 & 787\\
Theano & 2 & 58.33\% & 1114 & 653\\
OpenCV & 0 & 0\% & 0 & 480\\
Keras & 0.5 & 16.67\% & 211 & 211\\
Chainer & 3 & 75.00\% & 1465 & 1558\\
CNTK & 0.5 & 12.5\% & 0 & 92\\
MxNet & 1 & 14.29\% & 241 & 208\\ 
\textbf{Overall} & 1 & 33.33\% & 366 & 344 \\ 
\midrule
React & 1 & 25.00\% & 704 & 704\\
ETCD & 1 & 33.33\% & 704 & 1954\\
Flutter & 1 & 33.33\% & 483 & 809\\
Rancher & 3 & 66.67\% & 133 & 133\\
iPython & 2 & 66.67\% & 1649 & 1695\\
Babel & 2 & 42.22\% & 496 & 356\\
AWS-CLI & 1 & 25.00\% & 598 & 633\\
Drone & 1 & 25.00\% & 68 & 57\\
OSQuery & 1 & 33.33\% & 254 & 254\\
Grafana & 2 & 33.33\% & 1965 & 1965\\
\textbf{Overall} & 1 & 33.33\% & 754 & 704 \\ 
\bottomrule
\end{tabular} }
\end{center}
\vspace{-10pt}
\end{table}

Table~\ref{table:member1} shows the median values for each indicator for each system that we studied. Immediately, we see quite a bit of variance between systems in ML. Compared to traditional projects, ML projects vary wildly in terms of the percent of members that participate in issue discussion---from 12.50\% to 75.00\% of members taking part in discussions. In comparison, traditional frameworks show a narrower range of percentages, with medians of 25-66.67\% of members taking part in issue discussion. Overall, however, the median percentage of members taking part in issue discussion in ML and traditional frameworks are quite similar. % (medians of 33.33\%). 

There is quite a bit of variance in the number of contributions made by project members. Members of the PyTorch and Chainer communities contribute quite a lot each year, while members of the TensorFlow community contribute very little in comparison, possibly due to a larger community. There is quite a bit of variance for traditional frameworks as well. Overall, members of traditional projects make more contributions on a yearly basis. % than members of Software 2.0 projects---with medians of 754 in Software 1.0.
%\greg{Is this just an effect of the number of members? There are more TensorFlow members. It may be clearer if you normalize the number of contributions based on the number of members}.

Again, there is significant variance between individual systems in terms of the number of contributions made by the issue-fixing pull request creator during the year that each pull request was created. Compare with ML, pull request creators in traditional frameworks contribute more overall. %, with a median of 704 versus 344 for Software 2.0. 

%number of pull requests contributed each year. Interestingly, despite members contributing more overall to Software 2.0 systems, Software 1.0 systems receive more pull requests, with a median of \greg{fill in} versus \greg{fill in} for Software 2.0 systems. 

% \greg{how do the number of members compare to the number of pull request contributors? Do PR contributors tend to be members? What percentage of PR are made by non-members? I feel like a lot is missing here that  could be investigated further}

\definecolor{shadecolor}{rgb}{0.9,0.9,0.9} 
\begin{shaded}
\textbf{Summary}: There is little we can say categorically about community activity level for ML versus software frameworks. Overall, the two categories show similar levels of member participation. 
Software frameworks members contribute more to open source software.
However, there are significant differences between individual systems. 
\end{shaded}

\section{Threats to Validity}

\textbf{Internal Validity:} First, our study involves manual inspections on system-related issues in machine learning frameworks. These subjective steps can be biased due to interpretation of intent based on limited code comments and issue description. In order to reduce this threat, one author analyzed the issues separately and discussed inconsistent issues with a second author until an agreement was reached. Second, our study investigated 453 issues from Github for 10 machine learning frameworks and 10 traditional systems. It is not clear how much our findings can or will generalize beyond our dataset, especially considering the fact that machine learning systems are evolving rapidly. However, it is not easy to expand this dataset. First, the manual efforts required to analyze the issues were large. We could automate the labeling process, but it would then introduce noise in how we categorise issues. To collect and analyze the issues, we spent approximately 960 person-hours, leading to an average 2.11 person-hours per issue. However, we believe that this process lead to stable conclusions for this exploratory analysis.

\smallskip\noindent\textbf{External Validity:} The main threat to the external validity is generalisation beyond the considered frameworks and selected issues. We selected the frameworks based on their popularity. To make the issue taxonomy as comprehensive as possible, we labeled a large number of issues from GitHub until we reached saturation of the categories. Since both ML and traditional frameworks are rapidly changing, the observations and relative numbers may change for each corresponding category. However, due to a large sample set, we believe that it is unlikely that the answers to the research questions would be impacted by sampling additional systems.

\section{Actionable Recommendations}

Based on our findings, we can make several recommendations to the developers of ML frameworks, as well as systems that make use of these frameworks. First, our results indicate that incorrect memory allocation, memory leaks, multi-threading errors, and performance regression occur more commonly in ML frameworks. Increased dependence on hardware selection, like the GPU, can also lead to issues. Developers should plan for handling these types of issues. It would be reasonable to actively recruit or advertise for developers who specialize in areas such as memory management, concurrency, or software product lines. Recruitment of developers with expertise in these topics could lead to better development, and faster response when an issue occurs. 

We found that many ML frameworks developers identify as researchers or some combination of engineer and researcher. This has both positive and negative implications. Researchers have specialized knowledge in their area of focus. This can be utilized to great benefit in developing ML frameworks. By taking advantage of this expertise, frameworks can deliver sophisticated, highly effective features. At the same time, it is important that overall development of a framework can proceed without losing sight of the ``big picture''. Developers should not focus solely on their own areas of expertise and ignore features outside of their focus area. The overall architecture of a framework, as well as its usability, are extremely important and require consensus and conversation across the team as a whole. It is important that the development community of a project crafts compatible API designs, coding standards, and testing standards that are followed across the project, and that developers have some knowledge of how their work influences the system as a whole. 

We also found that the users of ML frameworks tend to provide more detailed issue descriptions than those of traditional systems, perhaps reflecting the complex, specialized nature of such systems. This can be good, as more information can help developers reproduce and correct issues more easily. However, more text does not necessarily imply a greater quantity of useful information. It is important that users be given structure and guidance when reporting issues. ML framework developers should make use of issue report templates to ensure that important information is provided by reporters. TensorFlow and PyTorch communities are using templates for reporting the issues. Past experience can be quite useful in helping users file reasonable reports. Detailed issue reports, filed for past issues, can be used to provide examples to users filing new issue reports. Well-crafted issue reports should be retained and pointed to in order to help ensure that relevant details are included in new reports. 

Finally, we found that some issues such as API mismatches or incorrect memory allocation required more time, more discussion, and a greater number of involved users to come to a conclusion on whether there was an issue or how to fix it. The most contentious issue types reflect an evolving field and an active community. This is not necessarily a negative finding. In fact, it can be quite positive---a healthy culture where developers share ideas, debate the merits of them, and come to a consensus on a solution will often lead to rapid, sustainable improvement to a framework. Development communities should encourage and expect debate. This requires, however, the creation of moderation standards within a community to keep discussion on-target and civil. 

\section{Related Work}

%In this section, we will introduce previous related study about the different kinds of bug study for different systems. All of these previous studies provide the experience and idea about how to learn from the various systems' bug report effectively.

Others have tried to investigate the differences between the two system paradigms from various perspectives.
A previous empirical study analyzed issue reports for three open source ML systems including Apache Mahout, Lucene, and OpenNLP~\cite{thung2012empirical}. Programs bugs developed in TensorFlow have also been studied empirically~\cite{zhang2018empirical}. However, these studies focused on particular frameworks (e.g., TensorFlow) and collected all types of program issues (not necessarily systems-related issues) to the extent that they concluded ``the small number of performance inefficiency issues suggests either performance issues rarely occur or these issues are difficult to detect." We mainly focus on system-related issues, and we found that performance regressions are actually common symptoms in machine learning systems. 
While many prior studies exist on understanding the nature of system-related issues in the traditional software stack ~\cite{gunawi2014bugs,yang2018not,bhattacharya2013empirical,abal201442,lu2008learning,xu2017system}, our study explores a wider range of systems and issues, and offers a detailed comparison of machine learning with traditional systems. 

The findings of studies on traditional software systems may not apply in ML for multiple reasons, including the fact that in the ML stack, programming is done differently than in the traditional software stack. For example, when the network fails in a handful of rare cases in ML, we do not correct those predictions by correcting the code. Rather, those predictions are fixed by including more labeled examples of those rare cases in order to regularize the learning process~\cite{k2017software2}. 

Differences between the two paradigms have been investigated from the perspective of software engineering practices as well. For example, a case study at Microsoft~\cite{amershi2019software} details differences of developing in the AI domain versus traditional application domains, and how team processes and practices change. They identified three distinguishable aspects of ML: (i)~data accumulating, massaging and cleaning is much more complex, (ii) model customization require very different skill sets, and (iii) components are more difficult to handle as distinct modules. The testing process is also different in ML~\cite{zhang2019machine,xie2011testing,pei2017deepxplore,tian2018deeptest,srisakaokul2018multiple,cheng2018manifesting,pham2019cradle,srisakaokul2018multiple,murphy2007approach,xie2011testing,dwarakanath2018identifying,nejadgholi2019study}. Prior studies have also identified unique technical debt concerns for machine learning systems~\cite{46555,43146}. 

Many prior studies have examined performance-related issues in traditional software~\cite{wang2018understanding,yan2017understanding,killian2010finding,liu2014characterizing,yang2018not,zaman2012qualitative,han2018perflearner}. Each of these has informed our issue classification process. Configuration-related issues are also a significant concern in our research. A number of studies have been conducted on performance-related issues in software, systems, and cloud that informed our approach~\cite{xu2015hey,yin2011empirical,xu2016early,leitner2016patterns,herbst2018quantifying}.

%\greg{This is a little barebones. If we have room, more should be said about other studies on ML systems. I suspect we won't have room}
% Han et al. provide a detailed study on performance bugs and how to address them using issue reports~\cite{han2018perflearner}.

%Dwarakanath et al. propose an approach to identify implementation bugs in machine learning-developed metamorphic relations for an application, and are able to detect 71 percent of the implementation bugs~\cite{dwarakanath2018identifying}

\section{Conclusion}

Frameworks offer services that can be used to build software. Issues in frameworks will impact the software built using those frameworks. ML systems differ from traditional systems in how they execute, how configurations are managed, how systems are tested, and how and where they are deployed. Naturally, the issues that manifest will differ as well---as will how communities of developers behave in correcting those issues. We have conducted a moderate-scale study contrasting the differences in the system-related issues between popular ML and traditional frameworks. Our findings offer a number of interesting observations, with implications for the development of ML frameworks and systems that make use of these frameworks. We hope that this exploratory study as well as the recommendations will offer assistance to the ``machine learning systems'' community forming the best practices for this new paradigm.

\bibliography{sysml}
\bibliographystyle{ACM-Reference-Format}

%%%%%%%%%%%%%%%%%%%%%%%%%%%%%%%%%%%%%%%%%%%%%%%%%%%%%%%%%%%%%%%%%%%%%%%%%%%%%%%
%%%%%%%%%%%%%%%%%%%%%%%%%%%%%%%%%%%%%%%%%%%%%%%%%%%%%%%%%%%%%%%%%%%%%%%%%%%%%%%
% SUPPLEMENTAL CONTENT AS APPENDIX AFTER REFERENCES
%%%%%%%%%%%%%%%%%%%%%%%%%%%%%%%%%%%%%%%%%%%%%%%%%%%%%%%%%%%%%%%%%%%%%%%%%%%%%%%
%%%%%%%%%%%%%%%%%%%%%%%%%%%%%%%%%%%%%%%%%%%%%%%%%%%%%%%%%%%%%%%%%%%%%%%%%%%%%%%

%%%% Greg - Commented out the code examples. I don't think they are actually helpful. 
% \iffalse

\newpage
\appendix
\section{Appendix}

%%%%%%%%%%%%%%%%%%%%%%%%%%%%%%%%%%%%%

\subsection{Patch Code for Section 2}
\label{appendix:a}

\noindent\textbf{Resource Usage (Pytorch 7680 - Patch)}
\newcommand{\lstbg}[3][0pt]{{\fboxsep#1\colorbox{#2}{\strut #3}}}
\lstdefinelanguage{diff}{
  basicstyle=\ttfamily\small,
  morecomment=[f][\lstbg{red!20}]-,
  morecomment=[f][\lstbg{green!20}]+,
  morecomment=[f][\textit]{@@},
  %morecomment=[f][\textit]{---},
  %morecomment=[f][\textit]{+++},
}
\begin{lstlisting}[language=diff]
@@ -7,6 +7,7 @@
aten/src/THCUNN/im2col.h
template <typename Dtype>
+ __launch_bounds__(CUDA_NUM_THREADS)
__global__ void im2col_kernel(const int n, 
            const Dtype* data_im,
            const int height, const int width,
            const int ksize_h, const int ksize_w,

\end{lstlisting}

\begin{lstlisting}[language=diff]
@@ -58,6 +59,7 @@ void im2col(cudaStream_t 
stream, const Dtype* data_im, const int chan-
nels, _consistency(self):
aten/src/THCUNN/im2col.h
template <typename Dtype, typename Acctype>
+ __launch_bounds__(CUDA_NUM_THREADS)
\end{lstlisting}

\noindent\textbf{Performance Regression (Keras 8381 - Patch)}

\begin{lstlisting}[language=diff]
@@ -22,7 +22,6 @@
keras/engine/training.py
from .. import metrics as metrics_module
from ..utils.generic_utils import Progbar
- from ..utils.layer_utils import count_params
\end{lstlisting}

\begin{lstlisting}[language=diff]
@@ -967,8 +966,8 @@ def _check_trainable_weights
_consistency(self):
keras/engine/training.py
- if (count_params(self.trainable_weights) !=
-     count_params(self._collected_trainable_
-     weights)):
+ if (len(self.trainable_weights) !=
+     len(self._collected_trainable_weights)):
 warnings.warn(UserWarning(
      'Discrepancy between trainable weights 
      and collected trainable'
      ' weights, did you set `model.trainable` 
      without calling'
\end{lstlisting}

\noindent\textbf{Memory Leak (Tensorflow 2942 - Patch)}

\begin{lstlisting}[language=diff]
@@ -616,7 +616,7 @@ def _feed_fn:
' to a larger type (e.g. int64).')
- np_val = np.array(subfeed_val, 
-                   dtype=subfeed_dtype)
+ np_val = np.asarray(subfeed_val, 
+                   dtype=subfeed_dtype)
\end{lstlisting}

\noindent\textbf{API (Tensorflow 17932 - Patch)}

\begin{lstlisting}[language=diff]
tensorflow/contrib/data/python/ops/grouping.py
@@ -140,9 +140,9 @@ def bucket_by_sequence_length:
' Try explicitly setting the type of the feed tensor'
' to a larger type (e.g. int64).')
- def element_to_bucket_id(element):
+ def element_to_bucket_id(*args):
- seq_length = element_length_func(element)
+ seq_length = element_length_func(*args)
boundaries = list(bucket_boundaries)
buckets_min = [np.iinfo(np.int32).min] 
              + boundaries
\end{lstlisting}

\noindent\textbf{Memory Allocation (Incubator-Mxnet 7000 - Patch)}

\begin{lstlisting}[language=diff]
tensorflow/contrib/data/python/ops/grouping.py
@@ -140,9 +140,9 @@
#if MXNET_USE_CUDA
+ CUDA_CALL(cudaGetDeviceCount(&num_gpu_device));
+ CHECK_GT(num_gpu_device, 0) << 
+          "GPU usage requires at least 1 GPU";
  ptr = new storage::GPUPooledStorageManager();
#else
  LOG(FATAL) << "Compile with USE_CUDA=1 to 
                 enable GPU usage";
\end{lstlisting}

\end{document}